\documentclass[fleqn,10pt]{wlscirep}
\usepackage[utf8]{inputenc}
\usepackage[T1]{fontenc}
\usepackage{tabularx}
\usepackage{bm}
\usepackage{subfig}
\usepackage{multirow}
\usepackage{soul}

\usepackage{algorithm}
\usepackage{algpseudocode}

\title{Toward Personalized Darts Training: A Data-Driven Framework Based on Skeleton-Based Biomechanical Analysis and Motion Modeling}

\author[1,4,+]{Zhantao Chen}
\author[2,4,5,+]{Dongyi He}
\author[3,4,*]{Jin Fang}
\author[4]{Xi Chen}
\author[4]{Yishuo Liu}
\author[4]{Xiaozhen Zhong}
\author[4,*]{Xuejun Hu}

\affil[1]{School of Computer Technology and Application, Qinghai University, Qinghai, 810016, China}
\affil[2]{School of Artificial Intelligence, Chongqing University of Technology, Chongqing, 400054, China}
\affil[3]{Faculty of Education, University of Macau, Taipa, 999078, Macau SAR, China }
\affil[4]{Xiaoping Technology Innovation Lab, Zhongshan Xiaolan Senior High School, Guangdong‌, 528415, China}
\affil[5]{Department of Language Science and Technology, The Hong Kong Polytechnic University, Hung Hom, 999077, Hong Kong SAR, China}

\affil[*]{Corresponding: foldkant888@gmail.com (J.F), 13560630467@163.com (X.H)}

\affil[+]{These authors contributed equally to this work}

\begin{abstract}
As competitive sports training becomes increasingly refined, intelligent and data-driven, traditional dart training models—relying primarily on coaches' experience and visual observation, are increasingly unable to meet the demands of high-precision, goal-oriented movements. These demands include identifying individual differences, analysing movement quality and providing timely feedback. While previous studies have revealed the critical roles of release parameters, local joint movements and coordination control in dart throwing, existing quantitative analysis paradigms largely remain confined to examining local variables, describing single-release metrics or using static template matching. This is insufficient to support personalised decision-making in real-world training scenarios. Evaluation logic centred on a single 'standard movement' often overlooks functional equivalence and beneficial variability in motor skills, making it difficult to balance biomechanical interpretability, individual adaptability and the practicality of training feedback. This paper therefore proposes a data-driven dart training assistance system based on skeletal motion analysis and machine learning. The aim is to construct a closed-loop training support framework spanning motion capture, feature modelling, and personalised feedback output. The study uses a Kinect 2.0 depth sensor and an optical camera to capture dart-throwing data in markerless conditions that closely resemble real-world training environments. The system extracts 18 highly interpretable kinematic features centred on four core biomechanical dimensions: three-link coordination, release velocity, multi-joint angular configuration and postural stability. Two key modules were then developed on this foundation: first, a personalised optimal throwing trajectory fitting model combining historical high-quality samples with the minimum jerk criterion; and second, a motion deviation identification and recommendation generation model based on z-scores and hierarchical diagnostic logic. A total of 2,396 throwing samples were collected from professional and non-professional athletes to establish a data foundation for dart motion analysis. Experimental results demonstrate that the proposed trajectory fitting method can generate smooth, personalised reference trajectories that conform to the natural laws of human movement. Case studies further indicate that the system can identify key technical deviations, such as insufficient trunk stability, abnormal elbow displacement and imbalanced velocity control, and provide targeted training recommendations. This paper is significant not only because it proposes a functional training assistance system, but also because it attempts to break away from the traditional static template-matching paradigm by shifting the focus of dart motion evaluation from 'how much it deviates from a uniform standard' to 'how much it deviates from the individual's optimal control range'. To a certain extent, this framework enhances the specificity of motion analysis to the individual and improves the interpretability of training feedback. It provides a methodological basis for a scientific, quantitative and closed-loop approach to darts training and can be used as a reference for personalised training support in other high-precision, target-oriented sports.
\end{abstract}

\begin{document}

\flushbottom
\maketitle
%
%
\thispagestyle{empty}

\section*{Introduction}
As competitive sports continue to evolve toward greater precision, intelligence, and data-driven methodologies, traditional training methods—which rely primarily on coaches' experience and visual observation—are no longer sufficient to meet the demands of high-level competition in movement analysis, identification of individual differences, and timely training feedback\cite{houtmeyers2021managing,ikegami2014watching,passfield2017mine}. Against this backdrop, sports biomechanics—as a key interdisciplinary field that bridges human anatomy, neuromuscular control, and the principles of kinematics and kinetics—can transform intuitive understandings of movement into measurable, comparable objective parameters, thereby providing a scientific foundation for evaluating athletic performance and optimizing techniques\cite{wallace2022sports,vigotsky2019}. Currently, biomechanical methods are widely applied across various sports. For example, in sprinting research, researchers analyze the relationship between acceleration capacity and speed performance using metrics such as ground reaction forces\cite{nagahara2025}; In upper-body-dominated sports such as baseball and golf, motion capture technology is used to analyze the sequence of force transmission within the kinetic chain and its relationship to joint loads and hitting/throwing performance\cite{aguinaldo2025,scarborough2021association}. These studies demonstrate that biomechanics can effectively reveal the mechanical mechanisms underlying competitive movements.

However, as the research paradigm has gradually shifted from "describing movement" to "optimizing training decisions," the limitations of traditional biomechanical methods have become increasingly apparent. First, although high-precision metrics obtained under laboratory conditions offer strong explanatory power, their high equipment costs, spatial constraints, and complex data processing limit their widespread application in daily training \cite{yung2022characteristics,glazier2017towards}. Second, single or low-dimensional biomechanical metrics often fail to fully capture the complex systemic characteristics inherent in real-world movement performance. High-level sports performance is not a simple sum of isolated variables, but rather a dynamic, coordinated behavior resulting from the combined influence of an individual's morphology, neural control, environmental constraints, and task objectives \cite{glazier2017towards,bolt2024ecological}. Therefore, relying solely on traditional biomechanical metrics for movement assessment tends to oversimplify complex motor control issues, thereby weakening the specificity and situational adaptability of training feedback.

To overcome these limitations, computer vision and machine learning technologies are increasingly becoming key technical pillars for sports training support \cite{naik2022comprehensive, mundt2025bridging, Chen2025}. Compared to traditional laboratory paradigms, these methods offer greater adaptability to real-world scenarios and enhanced data processing capabilities, enabling continuous analysis of human motion, object trajectories, and spatiotemporal coordination under conditions that more closely resemble actual training environments. For example, in tennis research, visual tracking technology has been used to analyze the ball's flight trajectory and athletes' footwork patterns in real time\cite{yang2023ball,naik2022comprehensive}; in highly complex sports such as gymnastics and figure skating, multi-view pose estimation and machine learning algorithms have been employed to identify rotational movements, estimate body axis stability, and extract key technical features \cite{tian2023multi, cust2019machine}. These advancements demonstrate that artificial intelligence technologies have not only expanded the boundaries of sports data acquisition but have also enabled training assistance systems to transition from "passive recording" to "active analysis" \cite{rezaei2022automated}.

However, while existing research has made significant progress in motion recognition and technical analysis, there remain several critical gaps remain in its deeper applications for training decision-making. First, the final judgments of many systems still rely on coaches or experts for interpretation and adjudication, failing to truly achieve a closed-loop process from data collection and feature modeling to the output of training recommendations\cite{hecksteden2023why,naik2022comprehensive}. Second, existing movement evaluation methods remain highly dependent on the "template matching" approach, which compares the similarity between a tester's movements and a predefined "ideal movement". Related studies have attempted to score or identify technical movement sequences using methods such as dynamic time warping (DTW) \cite{halilaj2018machine}, but this approach implies a premise that warrants caution: the assumption that different individuals should all converge toward a single optimal template. In fact, athletic performance often exhibits functional equivalence and beneficial variability; athletes with different body types, strength levels, and control strategies may achieve similar or even superior task outcomes through different movement organization patterns\cite{bartlett2007movement,colyer2018review}. Therefore, if individual differences, developmental stage differences, and task-specific constraints are ignored, a single-template-oriented evaluation system may not only reduce diagnostic validity but also misguide the direction of technical adjustments, making it difficult to provide athletes with truly personalized and evolving training pathways.

Against this backdrop, darts provides an ideal setting for researching the quantitative analysis and personalized modeling of fine motor coordination. Darts is a high-precision, target-oriented sport that centers on hitting specific targets and places high demands on fine motor control of the upper limbs, postural stability, and sustained attention \cite{Nguyen_Pham_Pathirana_Babazadeh_Page_Seneviratne_2018}. Although it originated as a traditional recreational activity, the sport has gradually evolved into a professional discipline with a mature commercial infrastructure, widespread public participation, and well-defined competitive rules \cite{davis2024stand,rezzoug2018contribution,Davis_2018}. Existing research indicates that darts training plays a positive role in enhancing upper limb fine motor control, balance, and concentration\cite{scaramuzzi2025darts,song2025influence,whitehead2025searching,greve2023elite}; simultaneously, the "dart-throwing motion", a classic movement plane in wrist joint function research, continues to attract attention in orthopedics and rehabilitation medicine \cite{Nguyen_Pham_Pathirana_Babazadeh_Page_Seneviratne_2018,wolfe2006dart,lee2014}. More importantly, although the dart-throwing motion appears superficially as a brief, single-phase release of the upper limb, it essentially involves multi-joint dynamic coupling among lower-limb support, trunk stability, shoulder-elbow-wrist coordination, and control of terminal release parameters, making it a typical model for studying the fine-tuning mechanisms of the human kinetic chain \cite{robalino2025kinetic,rezzoug2018contribution,kudo2000}. Therefore, using darts as a starting point not only advances the scientific training of the sport itself but also holds promise for providing a reference framework for the general analysis of high-precision goal-directed movements.

Nevertheless, the development of current dart training support systems still lags significantly behind that of other more established competitive sports. This lag is not merely a matter of insufficient application of technical tools; more fundamentally, it reflects the lack of a stable and effective integration among the data foundation, evaluation paradigms, and feedback mechanisms. At the data level, the darts field has long lacked specialized, high-quality datasets of human joint motion suitable for quantitative modeling. Consequently, existing research is often constrained by sample size, data collection conditions, and experimental settings, making it difficult to support robust analysis of movement mechanisms or the generalization of models in real-world training environments\cite{rezzoug2018contribution}. Regarding motion analysis, dart throwing is a typical short-duration, high-precision, goal-oriented movement, with critical technical differences typically concentrated within a time window of less than a few hundred milliseconds. Relying solely on a coach's experience or conventional video playback often makes it difficult to consistently identify technical deviations that are subtle in magnitude yet have a decisive impact on the landing point\cite{walsh2011biomechanical,walsh2011capturing}. More importantly, most existing darts research still focuses on local parameters, single-release variables, or fixed coordination metrics, such as characterizing throwing control levels through release parameter compensation relationships\cite{kudo2000}. While such studies have indeed revealed several important mechanisms in darts motion control, their analytical paradigms generally remain biased toward mechanistic explanations themselves and have not yet been fully translated into a systematic feedback framework that can inform individual training decisions. Consequently, current dart training research still faces a structural tension: on the one hand, there is a lack of high-quality kinematic data sufficient to support continuous modeling and longitudinal comparisons; on the other hand, there is a lack of personalized evaluation systems that can balance biomechanical interpretability with machine learning adaptability. This structural limitation—characterized by "insufficient data support, static evaluation frameworks, and coarse feedback granularity"—is the key reason why dart training assistance systems struggle to achieve further refinement and practical implementation.

In light of the aforementioned issues, this paper proposes a data-driven darts training assistance system based on skeletal motion analysis and machine learning, aiming to establish a closed-loop training support framework that encompasses motion capture, feature modeling, reference generation, and feedback output. Addressing long-standing bottlenecks in darts training—such as data scarcity, static evaluation, and subjective feedback—this paper first establishes a biomechanical analysis framework for dart throwing based on four dimensions: three-link coordination, release velocity, multi-joint angle configuration, and postural stability. Based on this framework, 18 kinematic features with strong interpretability are extracted; Subsequently, by integrating Kinect 2.0 depth sensors with optical cameras, the study collected and processed 2,396 sets of throwing samples under markerless conditions that closely mimic real-world training environments, thereby establishing a data foundation for dart-throwing motion analysis. Building on this foundation, this paper further designed two core modules for training applications: first, an individualized reference trajectory fitting method that combines historical high-quality samples with the minimum-jerk criterion to generate smooth throwing references that conform to natural human movement patterns; second, a method for motion deviation identification and recommendation generation based on Z-scores and hierarchical diagnostic logic, which quantifies deviations of the current motion relative to the individual's stable performance range and outputs targeted training recommendations. Experimental results demonstrate that the proposed trajectory fitting method can generate individualized reference trajectories with sound biomechanical validity, while the diagnostic model can identify key technical deviations—such as insufficient trunk stability, abnormal elbow displacement, and imbalanced velocity control—and provide corresponding improvement recommendations. The significance of this paper lies not only in proposing a functional darts training support system, but also in attempting to break away from the traditional static template matching paradigm. It shifts the focus of motion evaluation from "how much the movement deviates from a uniform standard" to "how much it deviates from the individual's optimal control range." This provides a methodological basis for the scientific, quantitative, and individualized training of darts, and serves as a reference for training support in other high-precision, target-oriented sports. The main contributions of this paper are as follows:

\begin{itemize}

\item \textbf{Proposing a biomechanical analysis framework for dart throwing:} We construct a structured analysis framework for dart throwing based on four core dimensions—three-link coordination, release velocity, multi-joint angular configuration, and postural stability—to provide a theoretical foundation for subsequent feature extraction, result interpretation, and training feedback.

\item \textbf{Establishing a specialized dart-throwing motion dataset:} By integrating Kinect 2.0 depth sensors with optical cameras, we collected 2,396 throwing samples in a real-world, unmarked training environment, providing data support for quantitative analysis, individualized modeling, and training applications of dart-throwing motions.

\item \textbf{Designing a personalized reference trajectory fitting method:} Based on the screening high-quality historical samples and the minimum-jerk criterion, we generate reference trajectories that preserve individual characteristics while ensuring biomechanical validity, thereby overcoming the reliance of traditional static template matching on standardized motions.

\item \textbf{Implementation of a personalized feedback mechanism for training decisions:} Based on 18 kinematic features and Z-score-based diagnostic logic, this method quantifies deviations and identifies issues in individual throws, automatically generating targeted training recommendations to enhance the scientific rigor, interpretability, and practicality of dart training assistance.

\end{itemize}

\section*{Preliminaries}

\begin{figure}[h]
    \centering
    \includegraphics[width=0.8\linewidth]{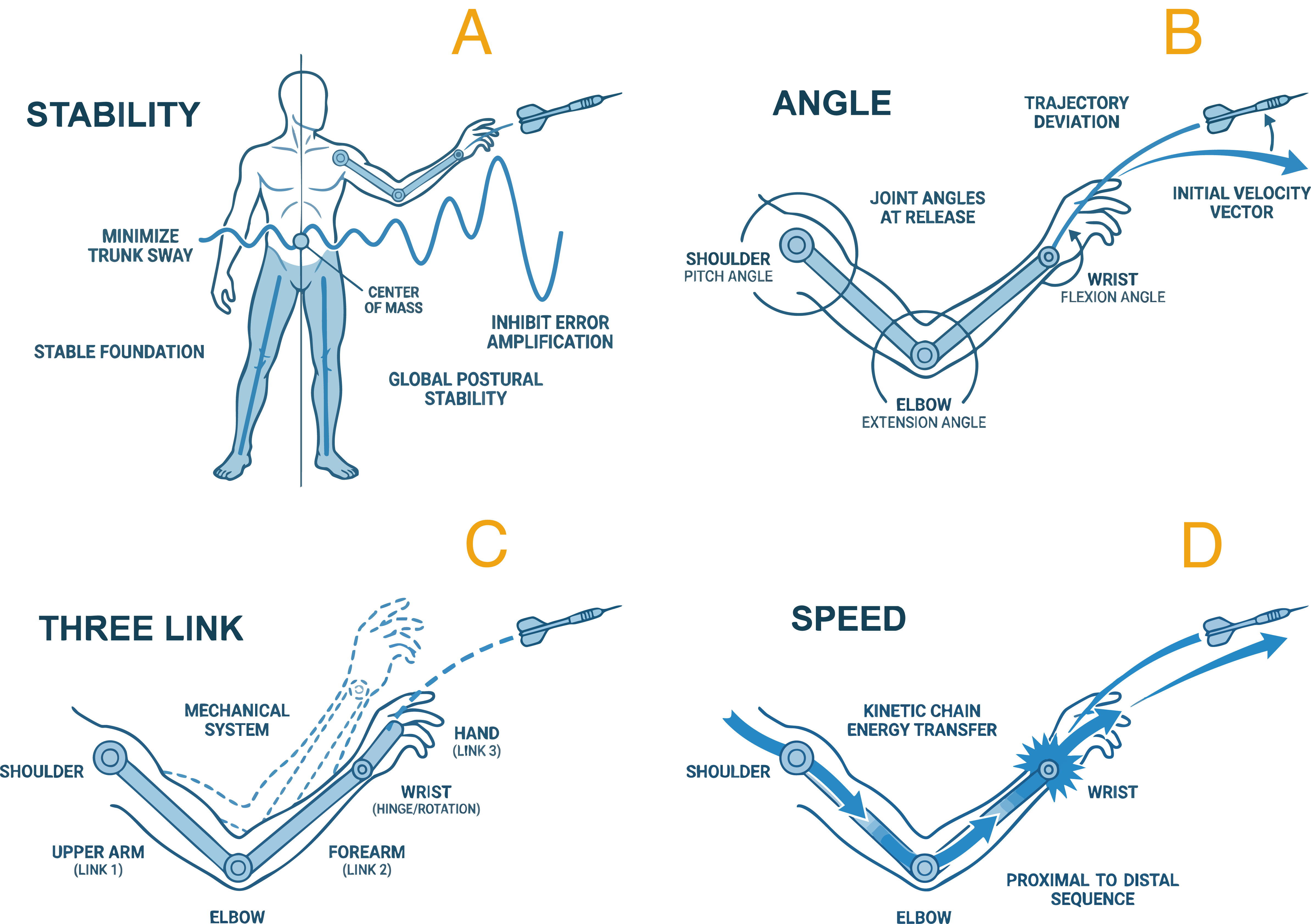}
    \caption{    
    \textbf{Four Factors Affecting Darts Throwing}
    (A) Postural Stability: Demonstrates the supporting role of global postural control. By establishing a stable foundation and minimizing the sway of the body's center of mass and trunk, it effectively suppresses the error amplification effect transmitted from lower-body disturbances to the upper-body kinetic chain. (B) Joint Angles: This section explains the spatial characteristics of the shoulder joint's pitch angle, elbow joint's extension angle, and wrist joint's flexion angle at the moment of release. The angular configuration of these key joints directly determines the direction of the dart's initial velocity vector; even the slightest angular disturbance can cause significant deviations in the flight trajectory. (C) Three-Link Motion Model: This model abstracts the throwing arm as a mechanical system composed of the upper arm (Link 1), forearm (Link 2), and hand (Link 3). It intuitively illustrates the physical framework and spatial coordination of the multi-link mechanisms—including the shoulder, elbow, and wrist—throughout the complete throwing cycle. (D) Release Velocity: This reveals the progressive energy transfer pattern based on the kinetic chain. Momentum strictly follows a "proximal-to-distal sequence," starting from the shoulder, transmitting through the elbow joint to the wrist joint, thereby achieving the most efficient power output at the moment of release at the distal end.}
    \label{fig:factor}
\end{figure}

Although dart throwing appears externally as a brief, discrete release movement of the upper limbs, its underlying mechanism cannot be reduced to the independent movement of a single joint or the linear output of a single instantaneous parameter. In terms of the essence of motor control, this movement is a complex process shaped by postural adjustment, segmental coordination, terminal release, and task constraints\cite{tran2019coordination,nasu2014two}. Therefore, before introducing motion evaluation models based on skeletal data and machine learning, it is necessary to first clarify the biomechanical factors that provide the core explanatory power in dart throwing and further discern the applicability limits of several commonly used technical indicators in practical training analysis\cite{panero2022biomechanical}. Combining existing research with the experience of professional coaches, this paper argues that dart-throwing performance can primarily be understood through four mutually coupled dimensions: the coordinated organization of the throwing arm as a multi-link system; the generation and transmission of release velocity; the orienting effect of multi-joint angular configurations on the terminal trajectory; and the supportive role of whole-body postural stability in fine upper-limb control, as shown in Fig.\ref{fig:factor}.

In the technical analysis of darts, the three-link kinematic model serves as a crucial theoretical framework for understanding the throwing motion. Relevant studies typically abstract the upper arm, forearm, and hand into a three-link system connected by the shoulder, elbow, and wrist, and employ multi-body kinematic methods to analyze the generation process of the terminal velocity of the projectile\cite{herring1992effects}. The theoretical value of this framework lies in its ability to transform the inherently continuous, high-degree-of-freedom throwing motion—which is difficult to compare directly—into a series of segmental rotational relationships with clear mechanical significance, thereby providing a structured descriptive pathway for motion analysis. Specifically, the shoulder joint typically serves as the proximal pivot and reference coordinate system; the elbow joint remains relatively stable during the back swing phase but undergoes spatial displacement during the forward swing phase to extend the terminal acceleration path and optimize the directionality of the release trajectory\cite{ambike2013leading,kudo2000}. However, the validity of this model rests on several idealized assumptions, such as segmental rigid-body behavior, simplified joint rotations, and the assumption that the motion occurs primarily within a local plane. For dart throwing in real-world training scenarios, trunk fine-tuning, shoulder girdle stability, individual morphological differences, and neural control strategies may all fall outside the explanatory scope of the three-linkage model. Consequently, the three-link model is better suited as a theoretical framework for organizing features and understanding synergistic relationships, rather than being directly regarded as sufficient evidence for judging the quality of a movement.

In conjunction with the structural model, release velocity and its formation process constitute another key mechanism underlying dart-throwing performance. Previous studies have shown that momentum transfer during dart throwing typically follows a characteristic "proximal-to-distal sequence," wherein energy and angular velocity originate at the shoulder, are transmitted through the elbow joint, and reach higher levels at the wrist and hand \cite{tillaar2019effect,rezzoug2018contribution}. The significance of this kinetic chain pattern extends beyond merely increasing dart release velocity; it lies in revealing the quality of the temporal organization of multi-joint synergistic force generation. Within the extremely brief time window before and after release, rapid wrist flexion and its coordination with elbow extension directly influence the dart's initial velocity, entry posture, and the consistency of its flight trajectory \cite{nasu2014two,james2018}. However, from a training and evaluation perspective, it is not rigorous to simply interpret release speed as "the higher, the better." On the one hand, the effect of speed is always constrained by the combined factors of release angle, equipment orientation, and release timing; on the other hand, the advantage of high-level athletes may not necessarily manifest as greater absolute output, but more likely as higher temporal stability and lower unnecessary fluctuations. Based on this, release velocity should be understood as an external manifestation of the outcome of coordinated control, rather than a single performance metric that can be interpreted independently of movement structure. For subsequent data-driven models, what truly holds analytical value is not merely the magnitude of velocity, but rather the coordination patterns and control quality inherent in the process of velocity generation.

Beyond the force output at the extremity, the configuration of multi-joint angles more directly influences the spatial orientation of the dart's trajectory at the moment of release. The angles of the shoulder, elbow, and wrist joints, along with their temporal evolution, collectively determine the hand's path, the dart's orientation, and the direction of the velocity vector at the instant of release \cite{rezzoug2018contribution,nasu2014two}. Previous studies have shown that elite athletes typically exhibit higher movement repeatability during elbow extension and wrist flexion/extension; this repeatability does not imply complete rigidity but rather reflects the relative stability of key functional parameters across consecutive throws \cite{tran2019coordination}. At the same time, minute angular perturbations at the moment of release can be amplified throughout the flight trajectory and result in significant deviations in landing points \cite{zhang2018exploiting}. Nevertheless, caution is warranted when interpreting "critical angles." If movement evaluation is based solely on a fixed angular template, it may overlook functional equivalence that exists between different individuals. In fact, athletes may well achieve similar release outcomes through different joint configurations and offset minor timing errors through local dynamic compensation\cite{nasu2014two}. Therefore, what is truly diagnostically significant is not a specific ideal angle value itself, but whether joint angles form a stable, repeatable, and outcome-effective control structure within the individual's own movement system. This also implies that subsequent feature modeling should not remain at the level of "how much the movement deviates from the standard template," but should further focus on the structural relationship between angular configuration and throwing outcomes.

In addition to the fine control of specific segments of the upper limbs, macro-level postural stability also constitutes a crucial foundation for high-precision dart throwing. Previous studies have shown that in goal-directed fine motor tasks, the degree of control over fluctuations in the body's center of mass, trunk sway, and disturbances in the base of support significantly affects the consistency of the trajectory at the upper limb execution end \cite{juras2013,zemkova2023sport}. In dart throwing, stable lower-limb support and trunk control can reduce mechanical noise in non-target directions, suppress errors transmitted from lower-body disturbances up the kinetic chain and amplified at the extremities, thereby provides more predictable mechanical conditions for dart release. However, equating postural stability simply with "keeping the body as still as possible" also has its limitations. Excessively rigid stabilization strategies may not only suppress necessary coordination adjustments but also increase the burden of local compensation, thereby diminishing the fluidity and adaptability of movement. A more reasonable understanding is that high-level performance relies not on absolute stillness, but on controlled stability—that is, the athlete's ability to constrain the body's degrees of freedom within a range advantageous to the task, while retaining sufficient room for fine-tuning to address temporal disturbances and minor errors. For data-driven analysis, this implies that trunk and overall postural characteristics should not be described solely in terms of minimizing displacement; rather, attention should focus on whether their fluctuation patterns align with stable hitting outcomes.

Overall, the technical performance of dart throwing cannot be reduced to a linear result of a single joint, a single moment, or a single template deviation, but should be viewed as a systemic product resulting from the combined action of multi-level biomechanical constraints \cite{furuki2017detecting}. The three-link model provides a necessary structured language for motion analysis, but its idealized assumptions mean it can serve only as a theoretical reference rather than a complete mechanism in itself; release velocity reveals the importance of terminal kinetic output, but its practical significance depends on deeper temporal organization and the quality of coordination; joint angles are direct variables influencing flight trajectory and landing point, but what truly warrants modeling is their functional stability, not static convergence to a fixed template; Postural stability provides the foundational support for fine-tuned control, but high-level performance relies not on absolute stillness, but on the controlled and effective management of fluctuations. Based on this understanding, the subsequent feature extraction and model design in this paper do not aim to identify a single "standard motion," but rather seek to strike a balance biomechanical interpretability and the preservation of individual differences, thereby providing a more robust theoretical foundation for personalized evaluation and dynamic feedback in dart training.

\section*{Methods}
To address long-standing issues in darts training—such as insufficient data support, static movement evaluation, and subjective feedback generation—this paper proposes a training assistance framework based on skeletal motion analysis and data-driven modeling. Rather than treating movement analysis, reference generation, and training feedback as isolated, independent components, this method aims to establish a closed-loop process encompassing data collection, feature representation, personalized reference modeling, deviation diagnosis, and recommendation generation. Overall, the method described in this paper consists of two interconnected phases.

This study first focuses on the characterization of kinematic features and the establishment of a data foundation. Specifically, this paper uses a Kinect 2.0 sensor to continuously collect 3D skeleton data during the dart-throwing process. By integrating this data with the biomechanical analysis framework proposed in previous research, key kinematic features are extracted from four dimensions: three-link coordination, release velocity, multi-joint angular configuration, and postural stability. The objective of this phase is not only to achieve a computable representation of dart-throwing motion but also to provide a feature set and data foundation with biomechanical interpretability for subsequent individualized modeling and deviation analysis.

Building upon the aforementioned feature representation and data foundation, this study further focuses on personalized modeling and feedback generation for training applications. Specifically, this paper designs two core modules: first, a personalized reference trajectory fitting model based on the selection of high-quality historical samples and the minimum-jerk criterion, which generates throwing references that balance smoothness with the preservation of individual characteristics; second, a motion deviation recognition model based on extracted kinematic features and Z-score–based diagnostic logic, which quantifies the degree of deviation of the current motion from the individual's stable performance range and automatically generates targeted training recommendations. By integrating these two stages, the method proposed in this paper achieves a complete training support workflow spanning from "motion capture and representation" to "reference establishment and feedback output."

\subsection*{Extracting Motion Characteristics of Dart Throwing based on Kinematic Data}

Based on the four core kinematic dimensions identified in the aforementioned theoretical framework, this study employed spatial vector algebra and time-series analysis to quantify the characteristics of athletes' throwing motions.
By extracting the three-dimensional spatial coordinates of each joint, this study ultimately established 12 transient kinematic features at the moment of release, along with six sequential continuous feature indicators covering the entire throwing cycle.

First, regarding the characteristics of dart velocity, this study focuses on the release speed and direction of the dart. By calculating the product of the three-dimensional displacement of the distal wrist joint between adjacent frames and the frame rate (time), this study derives the absolute linear velocity at the moment of release. The formula is as follows:
\begin{equation}
  Speed = \big\| \mathbf{p}_{\mathrm{hand}}(t) - \mathbf{p}_{\mathrm{hand}}(t-1) \big\| \cdot FPS,  \label{eq:release_speed}
\end{equation}
where $\mathbf{p}_{\mathrm{hand}}$ refers to joint 23 of the human skeleton, and $FPS$ denotes the video frame rate.

In addition, to quantify the spatial alignment of the throwing direction, the release alignment angle is calculated using the inverse cosine of the angle between the hand velocity vector and the target vector, according to the following formula:
\begin{equation}
  \theta(a,b) = \arccos \!\left( \dfrac{a \cdot b}{\|a\|\,\|b\|} \right), \label{eq:release_alignment}  
\end{equation}
where $a$ is the direction vector of the throwing arm, and $b$ is the vertical upward direction of the body.

At the dynamic level, time-series curves of hand velocity are continuously calculated and extracted.
The hand velocity sequence $s_{\mathrm{hand}}(t)$ is obtained by computing the first-order difference of the three-dimensional hand coordinates $\mathbf{p}_{\mathrm{hand}}(t)$ from adjacent frames, using the following formula:
\begin{equation}
  s_{\mathrm{hand}}(t) = \big\| \mathbf{p}_{\mathrm{hand}}(t) - \mathbf{p}_{\mathrm{hand}}(t-1) \big\|_2 \cdot \mathrm{FPS},  \label{eq:hand_speed_time_series}
\end{equation}
where $\mathbf{p}_{\mathrm{hand}}$ is the 23rd joint of the human skeleton, and $\mathrm{FPS}$ is the video frame rate.

Next, to quantify angular changes across multiple joints, this study constructed direction vectors using the three-dimensional spatial coordinates of adjacent limb segments (such as the upper arm, forearm, and hand). By applying a geometric method that calculates the angle of deviation through vector dot products, the study precisely determined static indicators such as the shoulder joint pitch angle, elbow flexion angle, wrist extension angle, and trunk yaw angle at the moment of release.
Regarding dynamic characteristics, these discrete spatial angles were further extrapolated along the time axis, and cubic spline interpolation was used to interpolate the data [48], with the formula is as follows:
\begin{equation}
  S_i(t) = a_i + b_i(t - t_i) + c_i(t - t_i)^2 + d_i(t - t_i)^3, \quad t \in [t_i, t_{i+1}],
\end{equation}
where $t_{i}$ denotes the joint angle at the $i$th frame, $a_{i}$ denotes the joint angle of the curve at time $t_{i}$, and $b_{i}$ denotes the slope of the curve at $t_{i}$, which physically represents angular velocity; $c_{i}$ is the angular acceleration of the curve at time $t_i$; and $d_{i}$ is the third derivative of the curve at time $t_i$, representing angular jerk.
By fitting discrete values using cubic spline interpolation, we construct dynamic trajectories of the angular positions of key joints—such as the shoulder, elbow, and wrist—over time, which are then used to analyze the angular momentum transfer mechanism during the throwing process.

Next, this study quantified stability metrics at both the micro and macro levels. For macro-level body stability, a specific time window centered on the release moment (five frames before and after release) was defined. By calculating the average displacement fluctuations of the spatial coordinates of the head, trunk, and ankle joints within this time window, the study developed static stability metrics for key body segments (head, trunk, and ankle stability), with the formula is as follows:
\begin{equation}
  Stability = \displaystyle \frac{1}{|S|-1}\sum_{t \in S} \big\|\mathbf{q}_t - \mathbf{q}_{t-1}\big\|,\quad S=\{r-5,\dots,r+5\} ,\label{eq:stability_macro}  
\end{equation}
where $S$ denotes the set of time windows, and $r$ denotes the release time of the dart; in this study, the release time is defined as the moment of maximum velocity. $\mathbf{q}_t$ represents the coordinates of the target joint (head, trunk, or ankle).

At the micro-level of grip stability, the degree of fine disturbance in finger control of the dart was rigorously quantified by calculating the mean distance and standard deviation between the thumb and fingertip nodes (mean grip distance and grip distance variability). The specific formula is as follows:
\begin{equation}
  \overline{d} = \frac{1}{T}\sum_{t=1}^{T} d_t, \quad d_t = \big\| \mathbf{p}^{\mathrm{HANDTIP^{(R)}}}_t - \mathbf{p}^{\mathrm{THUMB^{(R)}}}_t \big\| \qquad \qquad \sigma_d = \sqrt{ \frac{1}{T} \sum_{t=1}^{T} \big( d_t - \overline{d} \big)^2 },
\end{equation}
where \( HANDTIP^{(R)} \) denotes the tip of the thumb, \( THUMB^{(R)} \) denotes the base of the thumb, \( d_t \) represents the distance between the two at frame \( t \), \( \overline{d} \) is the average distance, and \( \sigma_d \) is the standard deviation.

Finally, based on the three-bar model, this study maps the release moment within the time frame to a percentage of the entire throwing cycle, thereby constructing a global temporal feature known as the release phase proportion, denoted as \(\tau_{rel}\), which is calculated as follows:
\begin{equation}
  \tau_{rel} = \frac{ridx}{T} \times 100\%,
\end{equation}
where $ridx$ is the frame index at the moment of release, and $T$ is the total number of frames in the entire throwing cycle.

Overall, based on the four primary factors influencing dart movement—velocity, angle, stability, and the three-link model—this study established a multidimensional feature engineering framework. It accurately extracted 12 static metrics at key moments and 6 dynamic features throughout the entire throwing cycle from raw 3D coordinates, thereby providing a biomechanically interpretable data foundation for subsequent motion analysis and the development of training recommendations.

\begin{table}[h]
  \centering
  \caption{Participant information and data collection summary}
    \begin{tabular}{lccccccc}
    \toprule
    \multirow{2}{*}{\textbf{Athlete}} & \multicolumn{4}{c}{\textbf{Professional athletes}} & \multicolumn{3}{c}{\textbf{Non-professional athletes}} \\
    \cmidrule(lr){2-5} \cmidrule(lr){6-8}
     & S1 & S2 & S3 & S4 & S5 & S6 & S7 \\
    \midrule
    Gender   & Female & Female & Male   & Male   & Male   & Male   & Male \\
    Height   & 170 cm & 168 cm & 168 cm & 175 cm & 176 cm & 173 cm & 170 cm \\
    Weight   & 58 kg  & 57 kg  & 55 kg  & 70 kg  & 55 kg  & 55 kg  & 60 kg \\
    Arm Span & 165 cm & 170 cm & 172 cm & 180 cm & 173 cm & 165 cm & 168 cm \\
    Num.     & 237    & 251    & 255    & 305    & 696    & 430    & 222 \\
    \bottomrule
    \end{tabular}%
  \label{tab:addlabe3}%
\end{table}%

\subsection*{Participants and Experimental Data Acquisition based on Kinect 2.0 Depth Camera and Optical Camera}
The study protocol was reviewed and approved by Zhongshan Xiaolan Senior High School, Zhongshan, Guangdong, China. All participants were minor high school students. Written informed consent was obtained from the legal guardians of all participants prior to data collection, and assent was obtained from the participants themselves. Participant-identifying information and research data were managed separately, and any potentially identifiable images or video-derived frames were anonymized prior to publication unless additional authorization had been obtained. All methods were performed in accordance with relevant guidelines and regulations for research involving human participants.

To minimize interference with the participants while improving the portability and reliability of the acquisition system, this study did not employ conventional wearable devices for motion capture. Instead, a Kinect 2.0 depth camera was used as the core acquisition device. The Kinect 2.0 provides high-resolution image and depth information, enabling accurate three-dimensional skeletal tracking during dart throwing under markerless conditions that closely resemble real training scenarios. In addition to joint motion data, an RGB camera and an intelligent dart machine were used to record dart landing positions, thereby enabling a more comprehensive assessment of movement performance.

A total of 2,396 data sets were collected from seven participants, including four professional dart players and three beginners; detailed participant information is provided in Table \ref{tab:addlabe3}. The resulting data set captures both the stable and standardized throwing patterns of professional players and the movement characteristics of students under non-professional training conditions, providing a robust basis for subsequent comparative analyses and model development.

\subsection*{Determining Dart Landing Points and Bullseye Positioning Using Tip Detection and Color Space Conversion}

\begin{figure}[hbtp]
    \centering
    \includegraphics[width=0.8\linewidth]{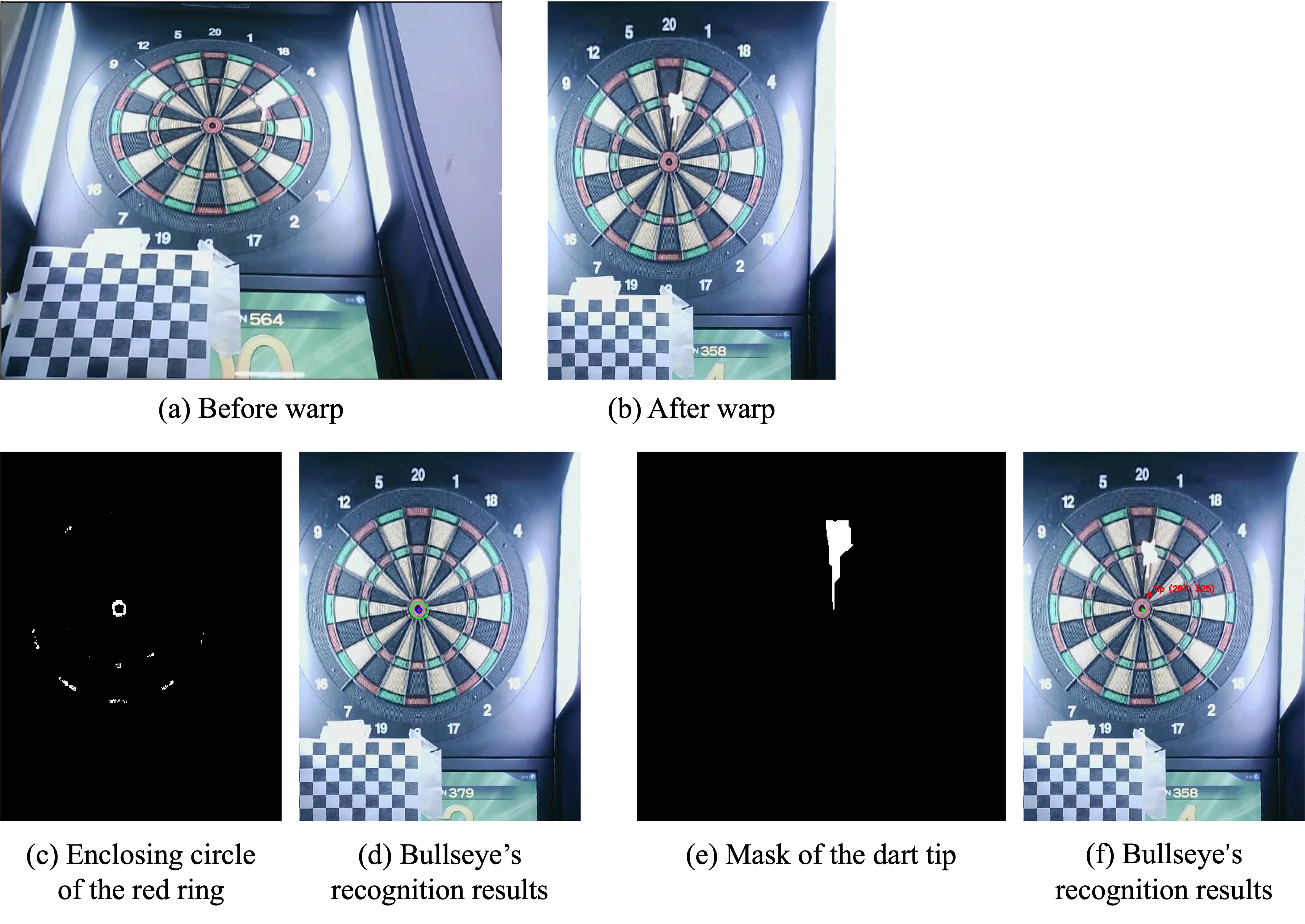}
    \caption{
    \textbf{Data preprocessing and dart-to-bullseye distance measurement pipeline.}
    Figure (a) displays the raw captured image exhibiting significant perspective distortion and tilt. Figure (b) shows the geometrically restored image following an inverse perspective projection using chessboard calibration parameters. Figure (c) illustrates the bullseye localization process, isolating the red outer ring via HSV color space conversion. Figure (d) presents the isolated dart tip mask generated through frame differencing and morphological operations. Finally, Figure (e) demonstrates the completed recognition result, accurately pinpointing both the bullseye center and the dart tip to compute the exact Euclidean landing distance.
    }
    \label{fig:darts}
\end{figure}

To obtain the landing coordinates of the darts, this study deployed an optical camera for data collection. However, placing the camera directly in front of the dartboard would inevitably interfere with the athlete's natural throwing trajectory, thereby affecting the authenticity and accuracy of the data. To avoid visual and physical interference with the test subjects, the camera was discreetly positioned below and in front of the dartboard. Additionally, to address the issue of image perspective distortion introduced by this oblique viewing angle, this study performed geometric calibration was performed on the images captured by the camera by placing a standard checkerboard calibration plate on the side of the target that lies in the same plane. The study first extracted the corner points of the checkerboard pattern in the image. Based on the topological distribution of the checkerboard, the corner positions of each sub-cell in the image were extracted as source coordinates. Combined with the actual physical dimensions of the checkerboard cells($2\times2 cm$), a target projection coordinate system was constructed. Based on the target mapping coordinates and image coordinates, the perspective transformation matrix of the image is solved. Finally, a perspective transformation was applied to the raw captured image to reproject it into a front view, thereby enabling precise extraction of the absolute positions of the dart's landing point and the bullseye.

Subsequently, this study proposes a feature detection method based on color space transformation to achieve target center recognition. Considering that red is primarily concentrated at the extremes of the first component in the HSV color space, this study employs a dual-threshold interval for joint segmentation. Based on experimental tuning, the parameters for the first interval are set to $H[0, 10]$, $S[60, 255]$ , $V[40, 255]$, while the parameters for the second interval were set to $H[168, 180]$, $S[40, 255]$, $V[160, 255]$. By filtering through these two threshold intervals, the initial mask of the red outer ring of the target disc was extracted. Simultaneously, HSV thresholding is applied to filter out the black central region, with specific parameters set to $H[50, 200]$, $S[50, 200]$, and $V[0, 70]$, yielding the initial mask for the black region. To remove noise from the mask, the system further applies a morphological closing operation using an elliptical structural element to the binarized image, effectively filling voids within the target region and smoothing edge noise. Finally, by calculating the minimum bounding circle of the red outer ring, the coordinates of the candidate target center and the outer radius $R_{out}$ are preliminarily determined. If multiple potential targets are present, and to filter out false targets and further improve recognition accuracy, this study combines the black region mask with the proportion $\tau$ of the black region relative to the total circular area to determine whether the candidate is a valid target. If $\tau$ is below the set minimum fill threshold of 35\%, the candidate center is deemed a false target and discarded, thereby completing the secondary verification of the target center.

\begin{algorithm}[H]\caption{Dart Tip and Bullseye Localization}
\label{alg:dart_detection}
\begin{algorithmic}
[1]\Require Video $\mathcal{V}$; Chessboard $(2\times2\text{cm})$\Ensure Dart tip $\mathbf{p}_{tip}$, Bullseye center $\mathbf{c}$\State Extract pre-throw $\mathbf{I}_1$ and post-throw $\mathbf{I}_2$ from $\mathcal{V}$;
\State $\mathbf{F}_1, \mathbf{F}_2 \leftarrow \text{warpPerspective}(\mathbf{I}_1, \mathbf{I}_2)$ using chessboard corners;

\State \textbf{// 1. Bullseye Detection}
\State Create red mask via dual-threshold HSV \& morphological closing;
\State Find min enclosing circle. If black region area $> 35\%$, confirm center $\mathbf{c}$;

\State \textbf{// 2. Dart Mask Extraction}
\State Compute absolute diff $\mathbf{D} \leftarrow |\mathbf{F}_2 - \mathbf{F}_1|$; Normalize $\mathbf{D}$;
\State Binarize $\mathbf{D}$ using $98\% \sim 100\%$ intensity threshold;
\State $\mathbf{M} \leftarrow \text{MorphologicalClose}(\mathbf{D}, \text{kernel}=2\times8\text{ ellipse}, \text{iter}=10)$;

\State \textbf{// 3. Tip Localization}
\State Extract largest contour $\mathcal{C}$ from $\mathbf{M}$; find its centroid $\bar{\mathbf{x}}$;
\State Filter $\mathcal{C}$ to keep candidate points $\mathcal{P}$ where distance to $\bar{\mathbf{x}}$ is $> 90\%$ of max distance;
\State Calculate curvature $\kappa$ for each point in $\mathcal{P}$;
\State $\mathbf{p}_{tip} \leftarrow \arg\max_{p \in \mathcal{P}} |\kappa(p)|$; 

\State \Return $\mathbf{p}_{tip}, \mathbf{c}$
\end{algorithmic}\end{algorithm}

To accurately extract the actual landing point of a dart, this study developed a tip-localization algorithm\ref{alg:dart_detection} based on the fusion of morphological and geometric features. To extract a preliminary template of the dart, this study uses two frames of images captured before and after the dart is thrown, calculates the absolute difference between the two frames, normalizes the result, and further extracts the preliminary template using a brightness threshold of $98\%\sim100\%$. Due to the dart's slender physical structure and motion blur, the difference mask is highly prone to fragmentation. Therefore, the system applies a 2×8 asymmetric elliptical structural element to perform 10 consecutive morphological closing operations on the image, thereby effectively connecting the fragmented main body of the dart and generating the final dart mask. After obtaining the dart mask, the algorithm filters out interference from false tips such as the tail of the dart and selects the edge points located farthest from the contour's center of mass, provided their distance exceeds $90\%$ of the contour's length. Finally, to determine the precise location of the dart tip, the study calculates the curvature of each edge point and searches for the point with the absolute peak curvature, thereby determining the final precise coordinates of the dart tip, as shown in Fig.\ref{fig:darts}.

\subsection*{Optimal Throw Motion Fitting based on Minimum-Jerk Principle}

Traditionally, darts players have relied primarily on coaches' experience and demonstrations to guide their throwing techniques. However, this approach makes it difficult for coaches to clearly convey their intentions, and players face significant challenges in grasping the key points of the throwing motion. Furthermore, there is currently no method for establishing standardized throwing motion templates tailored to the characteristics of individual dart players. Therefore, personalized standard dart-throwing motion templates based on the athlete's own data would significantly improve training efficiency. To this end, this study utilizes a large dataset of dart-throwing motion records to conduct standard motion fitting experiments, with the aim of obtaining personalized optimal trajectories and motions.

To achieve this goal, the study divides the process into two parts. In the first part, the study will select a subset of high-quality throwing trajectories from the athletes' historical data. These high-quality trajectories are selected based on a motion scoring scheme that considers both trajectory stability as well as the distance between the dart's landing point and the center of the target.

In the second part, the study fits the selected high-quality trajectories using the minimum-jerk model. The Minimum-jerk model is a classical kinematic model that assumes human movement tends to occur in a manner that minimizes the rate of change of acceleration, or "jerk." Finally, by experimentally optimizing the model parameters, this study developed a method capable of generating smooth, biomechanically optimal throwing trajectories, thereby providing athletes with a personalized motion reference template.

In terms of the selection method, the screening algorithm used in this study combines the distance between the dart's landing point and the bullseye with the rate of change in acceleration during the throw to generate a score, thereby selecting trajectories with higher scores.The specific formula for the motion evaluation score $score$ is as follows:
\begin{align}\label{key2}
  \begin{split}
      score = \frac{\exp(a \cdot \text{distance})}{\sqrt{1 + b \cdot \text{jerk}}}
      \qquad jerk \;=\; \sum_{k} \bigl\lVert \Delta^3 \mathbf{h}_k \bigr\rVert_2^2,
  \end{split}
\end{align}
where $distance$ represents the distance between the dart's landing point and the bullseye, and $jerk$ refers to the rate of change of acceleration. This study employs a combination of a penalty coefficient $b=1\times10^4$ and a distance weight $a=0.25$. Experimental validation has shown that this combination keeps the deviation in landing position and the assessment of biomechanical metrics within the same order of magnitude, thereby preventing the $score$ from being unduly influenced by any single metric.

In optimal curve fitting, this study employs a fitting method based on minimum acceleration changes to fit standard movements.

Specifically, the study selects the $Top-30$ trajectories with the most stable and accurate performance from the athlete's most recent 200 throws, based on the movement score $score$, as the data source for fitting trajectories.
Subsequently, the algorithm uses the right shoulder joint as a spatial anchor to perform a global alignment of the 3D coordinates for this subset of trajectories.
Building on this, the study introduces a dynamic weighted averaging strategy to synthesize a preliminary standard throwing template. Specifically, the spatial nodes of each reference trajectory $\mathcal{A}^{(i)}$ are weighted and aggregated according to their specific action weights $w_i(a)$, thereby generating the preliminary optimal throwing trajectory $\widehat{\mathcal{A}}^{\ast}$. The mathematical definition of this weighting mechanism is as follows:
\begin{align}\label{eq:weight}
  \widehat{\mathcal{A}}^{\ast} = \sum_{i=1}^{K} w_i(a) \mathcal{A}^{(i)}, \qquad
  w_i(a) \;=\; \frac{\exp\!\big(a^* \, distance^{i}\big)}{\sqrt{\,1 + b \cdot jerk^{i}\,}}, 
  \quad i \in \mathcal{S},
\end{align}
where $\mathcal{S}$ denotes the set of the athlete's throwing trajectories, $distance^{i}$ is the distance between the landing point of the $i$th dart throw and the target, and $jerk^{i}$ is the rate of change of acceleration minimized during the $i$th throw. It is worth noting that during the calculation of the weights $w_i$, the algorithm no longer uses a fixed weight parameter $a$, but instead treats the parameter $a^*$ as an adjustable hyperparameter for optimization.
The optimization objective is to minimize the mean squared error (MSE) between the weighted, preliminarily optimized throwing trajectory and the 30 actual trajectories, thereby identifying the optimal parameter $a^*$,
\begin{align}\label{eq:aopt}
  a^\ast \;=\; \arg\min_{a} \;\sum_{i \in \mathcal{S}}
  \text{MSE}\!\Big(Q(a),\, \widehat{\mathcal{A}}^{(i)}\Big),
\end{align}
where $Q(a)$ denotes the weighted average of the generated preliminary optimal throw trajectories, and $\widehat{\mathcal{A}}^{(i)}$ denotes the set of coordinate points of the $i$th actual trajectory.

To further improve the biomechanical accuracy of the standard throwing motion, this study performed an in-depth smoothing optimization of the initially synthesized optimal trajectory $\widetilde{\mathcal{A}^{*}}$.
The optimization objective function consists of two parts: first, a spatial fidelity term, which requires the final trajectory $\mathcal{A}^\ast$ to align as closely as possible with the averaged template; second, a kinematic smoothing term, aimed at minimizing the third-order difference of the trajectory along the time dimension to conform to the principle of minimal jerk in natural human movement. By constructing and rigorously solving the symmetric positive-definite matrix equations derived from this problem, the system ultimately outputs an optimal reference trajectory $\mathcal{A}^\ast$ that is highly smooth and consistent with human biomechanical characteristics.
Specifically, the aforementioned optimization objectives and their algebraic solution process can be formally expressed as:
\begin{align}\label{key10}
  \mathcal{A}^{\ast}
  \;=\;
  \arg\min_{\mathcal{A}}
  \;\frac{1}{2}\,\big\|\mathcal{A}-\widetilde{\mathcal{A}}^{*}\big\|_{F}^{2}
  \;+\;
  \frac{\lambda}{2}\,\big\|D_{3}\,\mathcal{A}\big\|_{F}^{2},
\end{align}
where $D_3$ represents the third-order derivative of the trajectory and, like $jerk$, indicates the acceleration; $\lambda$ is a regularization weight coefficient that balances spatial fidelity and trajectory smoothness. As experience show that $\lambda=5$ achieves the optimal balance between preserving the athlete's individual movement characteristics and eliminating high-frequency kinematic noise.

\begin{figure}
    \centering
    \includegraphics[width=\linewidth]{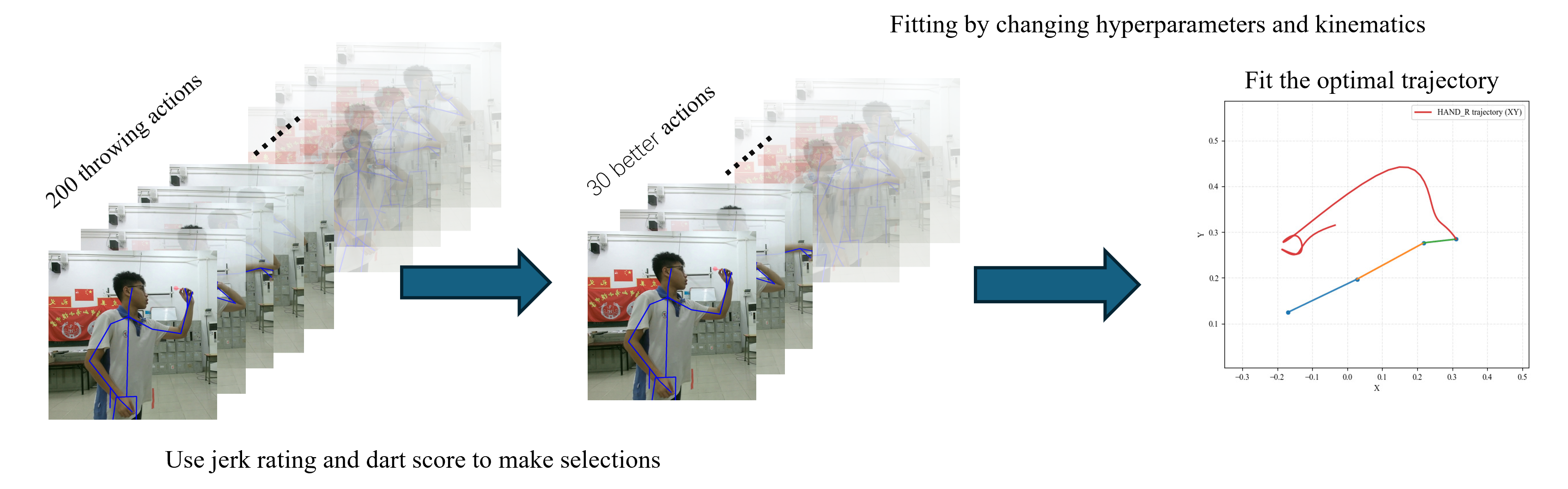}
    \caption{
    \textbf{Flow chart of optimal trajectory fitting.}
      This figure illustrates the data-driven pipeline for generating a personalized optimal dart-throwing trajectory. The process initiates by aggregating the athlete's most recent 200 throwing actions from historical data. Subsequently, a high-quality subset of 30 representative actions is extracted using a rigorous weighting scheme that evaluates both dart-to-bullseye accuracy and kinematic smoothness (jerk rating). Finally, these superior actions are synthesized via hyperparameter optimization and refined using a minimum-jerk kinematic model. This automated filtering and fitting mechanism ultimately yields a smooth, physiologically natural reference trajectory tailored to the individual athlete.
    }
    \label{fig:placeholder}
\end{figure}

In summary, the 30 most accurate and stable trajectories were first selected from the most recent 200 movements. By optimizing the hyperparameter $\alpha^{*}$, the model was trained to reproduce the trajectory with the lowest mean squared error among these 30 trajectories, and further optimization for minimum acceleration was applied to enhance smoothness.
The resulting optimal throwing trajectory not only preserves the athlete's individual movement control strategy but also conforms to the natural laws of human biomechanics.

\subsection*{A Method for Generating Training Recommendations for Athletes Based on Dart Characteristics}

To enable quantitative assessment of darts-throwing performance and provide personalized guidance, this study developed a diagnostic algorithm that integrates multidimensional darts-throwing motion features with an optimal reference model.
During the benchmark construction phase, the algorithm utilized the trajectory evaluation function described earlier to select the top 30 (Top-30) high-quality throwing trajectories with the highest scores.
Subsequently, 18 motion features were extracted from these high-quality trajectories, and the statistical mean and standard deviation of each feature variable were calculated. This established a data baseline for subsequent quantification of motion deviations and comparative analysis.

During the motion performance evaluation phase, 18 motion features of the current throw must be extracted.
To accurately measure local deviations between the current motion and the ideal motion pattern, this study introduces a $Z-score$ analysis mechanism, comparing the feature values extracted from the current throw with the statistical distribution of the optimal reference trajectory set. Regarding the absolute values of the calculated Z-scores $|z_j|$, to generate personalized improvement recommendations for each feature and alerts athletes to the shortcomings of their current movements, this study established a three-tier evaluation system to quantify the degree of movement deviation and defined corresponding performance intervals:
\begin{equation}
\operatorname{assessment}(z_j)=
\begin{cases}
\text{acceptable}, & |z_j| \le 1.0,\\
\text{slight deviation}, & 1.0 < |z_j| \le 2.0,\\
\text{significant deviation}, & |z_j| > 2.0.
\end{cases}
\end{equation}

\begin{figure}[htbp]
    \centering
    \includegraphics[width=\linewidth]{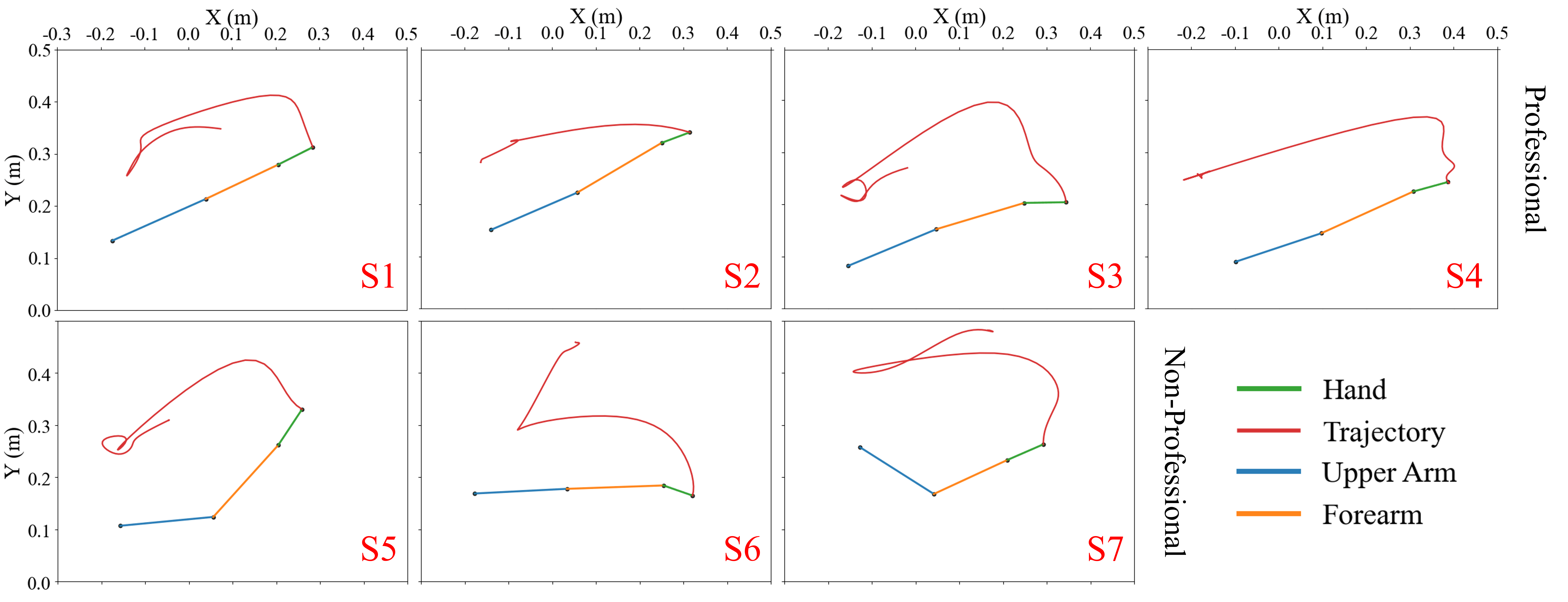}
    \caption{
    \textbf{Visualization of the optimal trajectories for seven dart players.}
    The figures in this experiment visualize the fitting results. The top four images show the fitting results for professional athletes, while the bottom three show those for non-professional athletes. You can observe each athlete's throwing process and hand movements. The green segments represent the motion from the wrist to the fingertips, orange segments represent the motion from the elbow to the wrist, and blue segments represent the motion from the shoulder to the elbow. These visualizations of the optimal fitting trajectories effectively illustrate the athletes' throwing processes, providing valuable references for darts training.}
    \label{fig:Curvers}
\end{figure}

\section*{Results}\label{sec4}

Based on the aforementioned methodological framework, this study has established a comprehensive analysis workflow ranging from motion capture of throwing actions, extraction of kinematic features, modeling of reference trajectories to the generation of personalized feedback. This section primarily reports results on two levels: first, an evaluation of the performance of the individualized reference trajectory fitting method based on the Minimum-jerk criterion in terms of trajectory smoothness, motion fidelity, and biomechanical plausibility; second, it presents the application results of a personalized diagnostic method—based on 18 darts-throwing kinematic features and a Z-score mechanism—in motion deviation identification and feedback generation. Overall, these two sets of results address the two key questions of "how to establish a reference" and "how to identify deviations," respectively, and together form the foundation of the results for the closed-loop training support framework presented in this paper.

\subsection*{Results of Individualized Reference Trajectory Fitting Using the Minimum-Jerk Principle}

In the personalized reference trajectory fitting experiment, the system first selects the top 30 high-quality trajectories from each athlete's most recent 200 throw records using a scoring function based on a combination of landing accuracy and trajectory smoothness; subsequently, personalized reference trajectories are generated through hyperparameter optimization and a minimum-jerk smoothing constraint. The goal of this process is not to replicate a single ideal template, but rather to extract representative kinematic references from historical high-performance data while preserving individual movement characteristics.

The fitting results indicate that this method can reliably generate continuous, smooth, and individually tailored reference trajectories for participants with varying skill levels, as shown in Fig.\ref{fig:Curvers}. For the four professional athletes and three non-professional athletes, the fitted hand trajectories all exhibited good spatiotemporal continuity, with no obvious high-frequency fluctuations or unnatural inflection points, indicating that the minimum-jerk constraint is effective in suppressing noise disturbances and maintains trajectory smoothness. At the same time, the trajectory patterns across different individuals did not become homogenized; instead, they retained their respective differences in movement rhythm, swing amplitude, and segmental organization. This indicates that while the method achieves smooth optimization, it does not simply eliminate the athletes' original individual control characteristics.

Furthermore, from the perspective of segmental coordination, the fitted trajectory exhibits a relatively clear pattern of proximal stability and distal acceleration across the shoulder–elbow–wrist segments. In Fig. \ref{fig:Curvers}, the blue, orange, and green segments correspond to the movement relationships from the shoulder to the elbow, the elbow to the wrist, and the wrist to the fingertips, respectively. It can be seen that the reference trajectory generally conforms to the kinetic chain organization pattern of progressive release from proximal to distal segments during dart throwing. In particular, for professional athletes, the fitted trajectories exhibit higher consistency and a more compact spatial distribution, reflecting the high repeatability of their movement patterns; in contrast, although non-professional athletes also generate smooth reference trajectories, their trajectory shapes exhibit more pronounced individual variations in terms of swing amplitude and segment coordination. These results indicate that the proposed method is not only applicable to populations with varying skill levels but also capable of capturing differences in the maturity of motor control within a unified modeling framework.

Overall, the individualized reference trajectory fitting method based on minimum-jerk is capable of extracting reference trajectories from a limited set of high-quality historical samples that combine smoothness, biomechanical plausibility, and the preservation of individual characteristics, thereby providing a stable intra-individual reference for subsequent motion deviation analysis.

\subsection*{Results of a Personalized Evaluation Method Based on the Kinematic Characteristics of Darts}

In the personalized Evaluation experiment, the system first conducted a quantitative analysis of the throwing motions of seven participants based on four core dimensions: release velocity, joint angle configuration, postural stability, and three-link coordination. It extracted 18 key kinematic features, as shown in Table \ref{tab:characteristics}. Among the participants, S1–S4 were professional dart players, while S5–S7 were non-professional. Overall, these features exhibited significant inter-individual differences, indicating that the constructed feature framework effectively captures multidimensional variations in dart-throwing motions related to velocity control, postural organization, segmental coordination, and stability.

From a static perspective, there are significant differences in the distribution of release velocity, release aiming angle, shoulder, elbow, and wrist joint angles, as well as in stability metrics for the head, trunk, and wrist among different participants. For example, some non-professional participants exhibit a wider range of variation in release velocity and aiming angle, whereas some professional athletes demonstrate greater consistency in trunk stability and joint angle configurations. These differences suggest that darts performance is not determined by a single parameter but rather results from the combined effects of multiple kinematic factors. Meanwhile, from a dynamic perspective, average hand velocity, average target alignment angle, and time-averaged angles at key joints can also distinguish the movement organization patterns of different participants, indicating that dynamic characteristics provide complementary value in characterizing the rhythmic structure and state of continuous control during the throwing process.

After establishing an individual baseline, the system compares the 18 features of a single throw with the statistical distribution of the individual's high-quality reference actions, quantifying the degree of deviation using Z-scores. The results demonstrate that this mechanism can identify the degree of abnormality across different motion dimensions on a unified scale, thereby transforming motion differences that were previously difficult to compare intuitively into interpretable quantitative results. Through this process, the system is not only able to determine whether a particular feature deviates from the individual's stable performance range but also performs a graded diagnosis based on the magnitude of the deviation, classifying motion status into three levels: acceptable, mild deviation, and significant deviation.

Furthermore, based on the Z-score-based deviation detection results, the system can map anomalies across different dimensions into training prompts with clear directionality, thereby achieving a closed-loop process of "feature extraction—deviation quantification—feedback output." In other words, the 18 features listed in Table\ref{tab:characteristics} not only constitute a set of descriptive metrics for dart-throwing motions but also form the quantitative foundation for personalized diagnostics. Consequently, the system can assess the degree of deviation of the current motion from the individual's stable performance range without relying on a single group template, and generate feedback tailored for training applications.

Overall, the personalized diagnostic method based on 18 kinematic features and a Z-score mechanism can systematically identify key movement deviations in dart throwing and provide quantitative evidence for subsequent training interventions. This indicates that the proposed framework not only establishes individualized benchmarks but also enables movement assessment and feedback generation based on these benchmarks, thereby forming a relatively comprehensive training support process.

\begin{table}[htbp]
  \centering
  \caption{Table of the 12 Static and 6 Dynamic Characteristics of Darts}
  \resizebox{\textwidth}{!}{
    \begin{tabular}{ccccccccc}
    \toprule
        & \multirow{2}{*}{\textbf{Characteristics}} & \multicolumn{4}{c}{\textbf{Professional athletes}} & \multicolumn{3}{c}{\textbf{Non-professional athletes}}\\
    \cmidrule(lr){3-6} \cmidrule(lr){7-9}
       
          & & S1    & S2    & S3    & S4    & S5    & S6    & S7 \\
    \midrule
    \multirow{12}{*}{\rotatebox[origin=c]{90}{Static}}
          & Release Velocity(m/s) & 5.57$\pm$0.58 & 5.31$\pm$0.72 & 5.15$\pm$0.57 & 5.90$\pm$0.81 & 4.67$\pm$0.64 & 6.80$\pm$0.91 & 7.20$\pm$0.75 \\
          & Release Alignment Angle($^{\circ}$) & 13.74$\pm$3.40 & 18.11$\pm$3.23 & 13.54$\pm$2.83 & 13.39$\pm$2.62 & 14.30$\pm$2.96 & 36.28$\pm$5.73 & 13.92$\pm$2.42 \\
          & Mean Grip Distance(mm) & 31.26$\pm$6.41 & 75.57$\pm$14.73 & 68.58$\pm$12.85 & 86.09$\pm$12.74 & 52.47$\pm$8.49 & 68.72$\pm$11.15 & 42.50$\pm$7.18 \\
          & Grip Distance Variability(mm) & 3.38$\pm$0.62 & 83.19$\pm$11.23 & 36.03$\pm$4.61 & 7.86$\pm$1.32 & 5.53$\pm$0.79 & 70.30$\pm$9.26 & 12.82$\pm$1.98 \\
          & Head Stability Index(mm) & 3.45$\pm$0.36 & 2.32$\pm$0.20 & 5.18$\pm$0.52 & 3.11$\pm$0.31 & 5.24$\pm$0.52 & 1.64$\pm$0.15 & 2.06$\pm$0.14 \\
          & Trunk Stability Index(mm) & 2.03$\pm$0.16 & 2.92$\pm$0.22 & 1.25$\pm$0.11 & 1.89$\pm$0.20 & 6.27$\pm$0.70 & 7.65$\pm$0.62 & 4.28$\pm$0.51 \\
          & Wrist Stability Index(m) & 0.51$\pm$0.05 & 0.63$\pm$0.06 & 0.65$\pm$0.06 & 0.66$\pm$0.04 & 0.52$\pm$0.04 & 0.77$\pm$0.06 & 0.59$\pm$0.06 \\
          & Shoulder Pitch at Release($^{\circ}$) & 69.66$\pm$10.50 & 80.77$\pm$14.19 & 61.63$\pm$9.12 & 58.03$\pm$12.48 & 71.73$\pm$13.44 & 48.56$\pm$9.27 & 57.94$\pm$10.93 \\
          & Elbow Flexion at Release($^{\circ}$) & 80.64$\pm$13.16 & 26.60$\pm$4.05 & 55.28$\pm$6.85 & 68.79$\pm$10.15 & 33.51$\pm$6.60 & 53.77$\pm$8.61 & 64.77$\pm$10.84 \\
          & Wrist Extension at Release($^{\circ}$) & 31.27$\pm$4.79 & 54.15$\pm$10.31 & 21.96$\pm$3.12 & 25.92$\pm$3.91 & 11.34$\pm$1.94 & 11.75$\pm$2.00 & 25.49$\pm$3.93 \\
          & Trunk Yaw at Release($^{\circ}$) & -13.94$\pm$1.04 & -8.81$\pm$0.79 & -12.78$\pm$1.23 & -15.16$\pm$1.26 & -20.95$\pm$1.35 & -15.88$\pm$1.05 & -21.83$\pm$1.65 \\
          & Release Phase Percentage(\%) & 83.02$\pm$12.77 & 92.38$\pm$17.22 & 76.46$\pm$15.67 & 87.57$\pm$16.66 & 90.23$\pm$19.94 & 83.35$\pm$13.16 & 77.38$\pm$12.22 \\
    \midrule
    \multirow{6}{*}{\rotatebox[origin=c]{90}{Dynamic}} 
          & Mean Hand Velocity(m) & 4.47$\pm$0.79 & 4.36$\pm$0.72 & 4.82$\pm$0.70 & 5.71$\pm$1.01 & 4.22$\pm$0.91 & 5.57$\pm$0.95 & 6.55$\pm$1.10 \\
          & Mean Target Alignment Angle($^{\circ}$) & 100.35$\pm$9.07 & 102.75$\pm$9.10 & 96.94$\pm$10.57 & 95.44$\pm$7.32 & 100.49$\pm$11.38 & 74.38$\pm$11.67 & 101.69$\pm$10.93 \\
          & Mean Shoulder Pitch($^{\circ}$) & 66.13$\pm$7.50 & 66.89$\pm$7.39 & 61.21$\pm$6.71 & 61.05$\pm$4.52 & 69.80$\pm$8.23 & 44.97$\pm$4.52 & 61.14$\pm$5.90 \\
          & Mean Elbow Flexion($^{\circ}$) & 73.95$\pm$6.83 & 51.02$\pm$4.71 & 56.35$\pm$6.86 & 63.60$\pm$6.07 & 48.79$\pm$3.87 & 56.52$\pm$4.84 & 74.14$\pm$7.99 \\
          & Mean Wrist Extension($^{\circ}$) & 15.93$\pm$1.83 & 25.14$\pm$3.04 & 29.69$\pm$2.86 & 32.47$\pm$3.86 & 17.39$\pm$1.82 & 23.42$\pm$2.78 & 10.91$\pm$1.47 \\
          & Mean Trunk Yaw($^{\circ}$) & -13.76$\pm$1.95 & -14.33$\pm$1.84 & -13.27$\pm$1.62 & -15.58$\pm$1.86 & -21.52$\pm$2.53 & -17.60$\pm$2.81 & -21.16$\pm$3.22 \\
    \bottomrule
    \end{tabular}%
    }
  \label{tab:characteristics}%
\end{table}%

\section*{Case Study}
The results presented above have demonstrated, at a general level, that the proposed framework is capable of simultaneously perform individualized reference trajectory modeling and motion deviation detection, thereby establishing a closed-loop training support process ranging from "reference establishment" to "feedback output." Building on this foundation, this section further selects representative individual cases to provide a more detailed analysis and explanation of the system's output results. Specifically, we first analyze the performance of the individualized reference trajectory generated using the Minimum-jerk method in terms of kinematic organization and biomechanical validity, using professional athlete S3 as an example. Subsequently, using non-professional participant S6 as an example, we demonstrate how the diagnostic model—based on 18 kinematic features and a Z-score mechanism—identifies key movement deviations and generates targeted training recommendations. Through these two cases, this paper aims to further illustrate, at the individual level, the practical application and interpretability of the framework across the three stages of "reference trajectory generation—motion deviation identification—training recommendation output."

\subsection*{Case Analysis of Optimal Throwing Motion Based on the Minimum-Jerk Principle}

To evaluate the validity of the optimal throwing motion generated based on the Minimum-Jerk Principle, this study selected athlete S3's trajectory and validated it across four core dimensions: velocity, angle, stability, and the three-link model. The analysis indicates that the fitted motion exhibits ideal kinematic characteristics.

Regarding dart release velocity, the throwing trajectory—represented by the index finger's key node—exhibited significant smoothness in its peak region. This smoothness extended the time window prior to release, allowing the athlete to have more ample time to aim at the moment of release, which plays a crucial role in improving hit rates. This finding aligns with the conclusions of Nasu et al., who noted that professional dart throwers typically possess a longer throwing window to enhance the success rate of their movements. Furthermore, the reference trajectory fitted in this study reaches a peak velocity of approximately $5.21 m/s$ at the moment of release. In actual throws, release velocity is a key variable controlling the distribution of dart landing points: too low a velocity causes the dart to land too low, while too high a velocity not only causes the dart to land too high but also excessively compresses the aiming time, thereby reducing overall accuracy. According to research by Venkadesan et al., the optimal release velocity range for darts is $5.1\sim5.5 m/s$. The velocity value fitted in this study perfectly aligns with this theoretical range, confirming its dynamical validity.

Regarding the release angle, this study analyzed the spatial dart-throwing posture corresponding to the fitted trajectory and measured the release angle to be approximately $17.5^\circ$. Because even minor deviations in the release angle are significantly amplified by the parabolic trajectory and affect the final landing point, precise control of the angle is crucial. Research by Venkadesan et al. similarly indicates that the optimal release angle for an overarm throwing strategy should fall between $17^\circ$ and $37^\circ$.

Regarding stability analysis, observation of the optimized throwing trajectory reveals that the joints of the hand are not in a state of absolute rest but exhibit slight spatial fluctuations within a reasonable range. However, these local joint adjustments do not disrupt the overall fluidity of the motion; the fitted trajectory remains highly smooth and stable, exhibiting an ideal parabolic shape during the dart's acceleration phase. The underlying mechanism of this phenomenon lies in the beneficial variability inherent in the human multi-joint system during movement. Although each joint possesses a certain degree of flexibility, these controlled micro-disturbances can compensate for one another through the kinetic chain, thereby ensuring the global stability of the overall throwing trajectory. This phenomenon is highly consistent with the findings of Kudo et al.; their research indicates that the "stability" of elite athletes is not a mechanical, absolute rigidity, but rather manifests in the ability to actively compensate for instability in the shoulder or elbow through fine-tuning across joints, thereby ensuring absolute consistency in the dart's release velocity and direction. \cite{kudo2000}.

Regarding the three-link kinematic model of dart throwing, the optimal throwing motion identified in this experiment closely aligns with the principles of momentum transfer within the human kinetic chain. Observation of the trajectory reveals that energy transfer in the optimal motion follows a "proximal-to-distal sequence": the shoulder and upper arm, serving as the proximal anchor points, initiate the force first; energy is then sequentially transmitted along the kinetic chain, driving the forearm to swing forward fully; finally, through rapid wrist flexion and downward pressure, the kinetic energy of the entire upper limb is efficiently converted into the dart's initial flight velocity. To ensure the stability of the entire momentum transfer process, the range of motion across the various segments of the upper limb exhibits significant spatial variation: the range of motion is smaller at proximal positions closer to the body's central axis, while it is larger at distal positions farther from the body. Specifically, during the entire forward acceleration phase, the elbow joint undergoes only a slight accompanying upward displacement toward the end of the throw, whereas the wrist joint maintains a large range of flexion and extension throughout. It is precisely this synergistic coupling of proximal stability and distal flexibility that enables the throwing motion to efficiently transfer kinetic energy while simultaneously controlling the dart's trajectory within a precise range, ensuring extremely high throwing stability. The trajectory characteristics fitted in this experiment are consistent with the findings of Kudo et al.; their study indicated that high-level dart throwers can perfectly utilize the momentum generated by the upper arm and forearm linkages to drive the third linkage (the hand), thereby significantly reducing the need for additional active force generation by the distal hand muscles and markedly improving throwing accuracy \cite{kudo2000}. Analysis of the throwing trajectory fitted for professional athlete S3 confirms that the fitting model developed in this study conforms to the objective laws of sports biomechanics and can provide athletes with supplementary guidance for their throwing movements.

\begin{figure}[htbp]
\centering
\includegraphics[width=\textwidth]{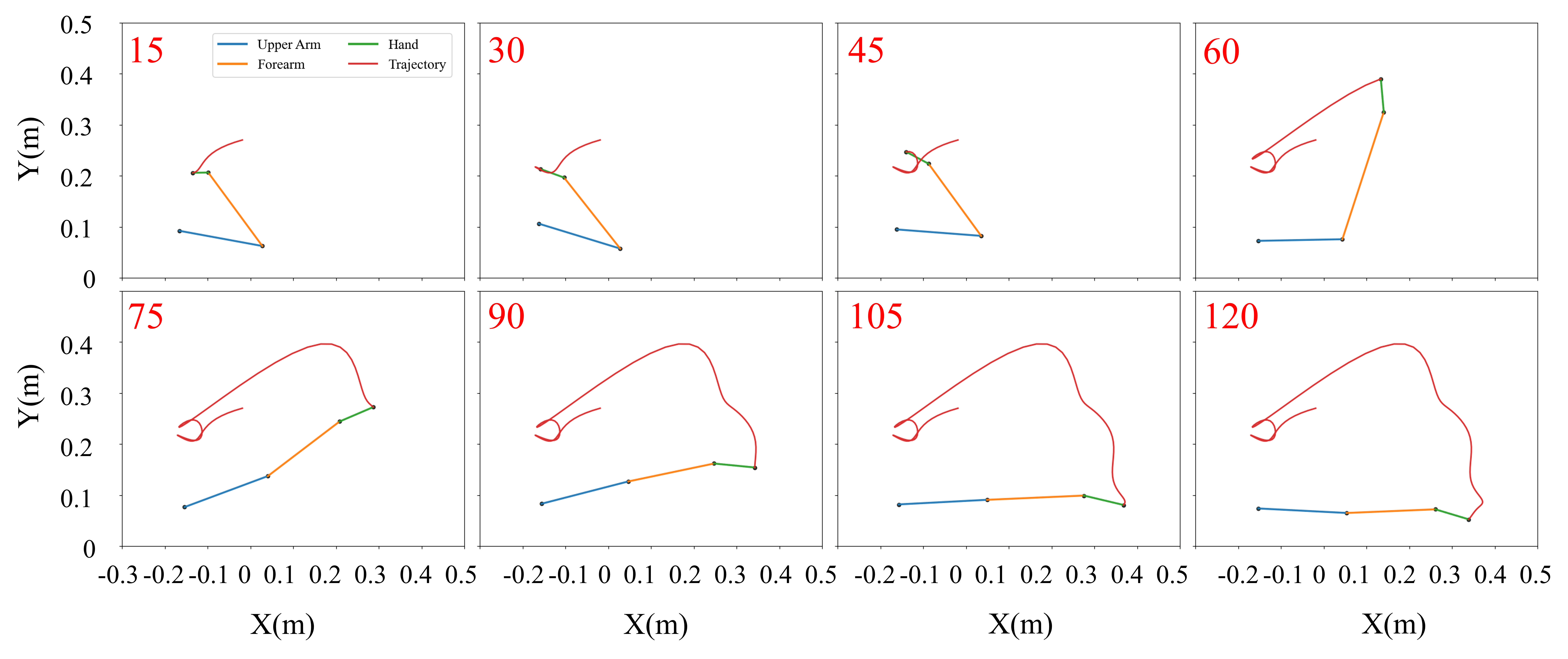}
\caption{
\textbf{Fitted optimal throwing trajectory for athlete S3.}
    This figure shows the time series of the personalized optimal throwing trajectory generated using the minimum jitter fitting method (from frame 15 to frame 120). The red curve depicts a smooth and continuous spatial path of the hand in the XY plane, indicating no excessive trembling or unnecessary fluctuations. Consistent with the principles of the three-link kinematic model, the athlete's shoulder remains highly stable throughout the entire movement. During the initial backward swing phase (frames 15 to 45), the elbow joint remains in a relatively fixed position to ensure mechanical stability. Subsequently, during the forward acceleration and release phases (frames 60 to 105), the elbow naturally rises to extend the acceleration path, ensuring efficient energy transfer.
}
\label{fig:the best curve}
\end{figure}

\subsection*{Case Study on a Method for Generating Training Recommendations for Athletes Based on Darts Performance Characteristics }

To make the evaluation results of the training recommendation generation method more convincing and to visually demonstrate the system's actual effectiveness in correcting technical deviations, this study selected a non-professional darts participant (designated S6) as a case study for in-depth analysis. S6 is a male with a height of 173 cm, a weight of 55 kg, and a wingspan of 165 cm. Compared to professional athletes whose movements are already highly established, diagnosing a beginner lacking systematic training allows for a more significant validation of the system's effectiveness in feature assessment and feedback generation.

\begin{figure}[hbtp]
    \centering
    \includegraphics[width=0.8\linewidth]{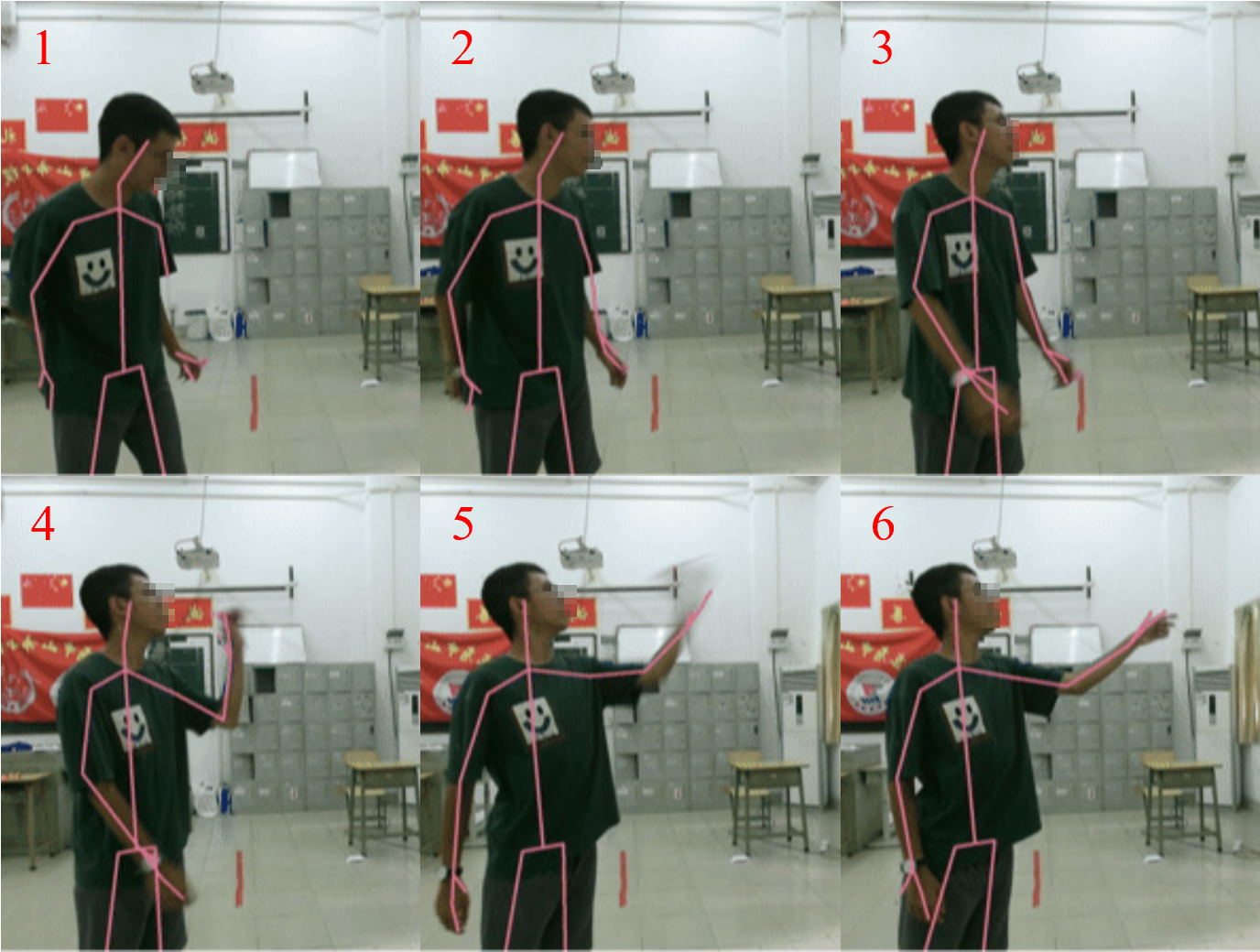}
    \caption{
    \textbf{Not standard action from S6.}
    This figure presents a non-standard throwing motion captured from a non-professional athlete across six sequential frames. The overlaid skeletal tracking highlights multiple kinematic deviations compared to an ideal reference trajectory. For instance, the sequence visually reveals inconsistencies such as noticeable trunk sway, inadequate head stability, and abnormal elbow displacement during the backward swing and acceleration phases. This specific motion data serves as the direct input for the personalized recommendation model. The system then calculates Z-score deviations across fundamental features to automatically generate targeted corrective feedback for the athlete.}

    \label{fig:Not_standard}
\end{figure}

After establishing a baseline, non-expert subject S6 underwent a new round of dart-throwing tests which shown as Fig.\ref{fig:Not_standard}. From a kinematic perspective, this subject’s throw exhibited three primary areas of non-conformity. First, the most significant issue was poor trunk stability. The subject leaned forward before the throw and then leaned backward during the power generation phase, resulting in noticeable displacement of the head and shoulders. This shifting of the center of gravity easily leads to deviations in the direction of force application, causing significant errors in the dart’s landing point relative to the target. Second, there were flaws in the “three-link” power generation mechanism of the hand. During the throw, the subject’s elbow joint exhibited significant movement, which violates the biomechanical characteristics of a standard throw. This results in disjointed force transmission, uneven dart acceleration, and a rushed release, directly causing deviations in the release angle. Finally, regarding speed control, the subject released the dart too quickly, shortening the effective time available to control the dart’s flight path. This tends to cause the dart to land too high and consequently reduces throwing accuracy.

In this case study, the training recommendation generation method proposed in this study effectively identified the movement deficiencies of subject S6 and provided a targeted improvement plan shown in Table \ref{tab:suggestion}. Specifically, to address S6’s poor trunk stability, the system generated guidance recommendations numbered 1–4, clearly identifying defects in head, trunk, and shoulder sway during dart release, and proposing corrective strategies such as “tighten the core” and “keep the center of gravity stable on the supporting foot.” Regarding the “three-link” force generation mechanism, the model accurately diagnosed abnormal displacement of the elbow joint and provided standard movement guidance, stating that “the elbow should remain largely stationary during the throw.” Finally, concerning speed control, in response to the issue of excessive release speed, the model recommended “reducing the amplitude of the backswing and maintaining a relaxed posture.” In summary, the recommendations generated in this study fully consider the key factors influencing dart throwing and are highly consistent with the principles of sports biomechanics. They effectively diagnose an athlete’s movement weaknesses and provide scientific training guidance.

\begin{table}[htbp]
  \centering
  \caption{Suggested sentences generated based on the characteristics of the S6 dart}
  \label{tab:suggestion}
  \begin{tabularx}{\textwidth}{c X}
    \toprule
    \textbf{No.} & \multicolumn{1}{c}{\textbf{Recommendation}} \\
    \midrule
    1 & Significant trunk instability and standard deviation during movement ($|z|_{\text{max}} = 11.55$): Focus on training movement consistency during the 100 ms prior to release. \\
    2 & Insufficient head stability during dart release ($z = 7.59$): Fix your gaze on the target, avoid nodding or side-to-side head movement, and maintain the same line of sight before and after release. \\
    3 & Insufficient trunk stability during dart release ($z = 6.92$): Engage your core, minimize body sway, and keep your center of gravity stable over your supporting foot. \\
    4 & Shoulder Forward Swing Angle ($|z|_{\text{max}} = 6.40$): Control arm elevation to align the upper arm’s trajectory with the target; avoid shrugging, which causes path deviation, and use the shoulder as a stable anchor point. \\
    5 & Abnormal elbow joint movement in the three-link mechanism ($|z|_{\text{max}} = 4.66$): The elbow should remain largely stationary during the backswing; only at a specific point during the forward swing—as the arm accelerates—should the elbow rise naturally, fully extending along a straight line upon release. \\
    6 & Acceleration curve is not smooth ($|z|_{\text{max}} = 4.65$): Maintain a continuous and even acceleration rhythm; avoid “stopping and then exerting force.” \\
    7 & Excessive release speed ($z = 1.60$): Simplify the backswing amplitude, relax the movement, and ensure that speed comes from smooth acceleration rather than a “violent flick.” \\
    8 & Aiming Angle Deviation ($z = 1.14$): Ensure the release path is straight, and avoid excessive wrist supination or pronation. \\
    \bottomrule
  \end{tabularx}
\end{table}

\section*{Discussion}
\subsection*{Key Findings and Research Significance}
This study addresses the core challenges in dart training—namely, insufficient data support, a static evaluation framework, and subjective feedback mechanisms—by developing a data-driven training assistance system based on skeletal motion analysis and machine learning. The results demonstrate that the system can perform structured modeling of dart-throwing motions across four biomechanical dimensions—multi-joint coordination, release velocity, joint angle configuration, and postural stability—under conditions of minimal intervention and in settings that closely mimic real-world training scenarios. Furthermore, it achieves a closed-loop analysis process spanning motion capture, feature extraction, reference trajectory generation, and the output of personalized recommendations. Compared to traditional observational training reliant on coaches’ experience, the value of this study lies not merely in “enhancing quantification,” but rather in its attempt to transform dart throwing—a high-precision, short-duration movement task characterized by significant individual variability—into a training decision-making problem that combines biomechanical interpretability with algorithmic operability.

From a theoretical perspective, a key contribution of this study lies in advancing the paradigm of dart-throwing motion analysis. Existing research on dart-throwing has largely focused on release parameters, local angles, or single coordination metrics. While these studies are significant for elucidating the mechanics of the movement, they often face two limitations in training applications: First, the lack of a unified organizational framework among these metrics makes it difficult to provide a systematic explanation of the entire movement; Second, research conclusions often stop at identifying “which variables are correlated with performance,” but rarely address the question of “how to translate these variables into actionable training feedback.” By integrating three-link coordination, release velocity, multi-joint angles, and postural stability into four mutually coupled analytical dimensions, this study essentially provides an explanatory framework that more closely approximates the actual motor control process. In other words, this study does not reduce dart-throwing motion to a single optimal angle or ideal velocity, but rather emphasizes that high-level performance stems from coordinated organization under multi-level constraints. This makes the analytical framework not only applicable to interpreting results but also provides a theoretical foundation for generating personalized feedback in future research.

Furthermore, this study seeks to overcome the methodological limitations of the traditional “template matching” paradigm. Many motion evaluation systems implicitly assume that all individuals should converge toward a single standard motion, but this assumption does not always hold true for goal-directed fine motor skills. Different athletes may achieve similar performance outcomes through distinct patterns of movement organization due to variations in body morphology, joint range of motion, force distribution, and control strategies. Based on athletes’ own historical high-quality throws, this study combines the Minimum-Jerk Principleto generate individualized reference trajectories. The core significance lies not in constructing a “universally optimal template,” but in establishing a reference frame that can be dynamically updated as the individual’s technical state evolves. Compared to static templates, this within-subject reference better aligns with the actual processes of skill learning and movement optimization, as the key to training lies not in mechanically replicating others’ movements, but in gradually converging toward the most effective and stable coordination structure for the individual.

The results of this study support this approach to some extent. Taking the S3 case as an example, the optimized trajectory obtained through fitting exhibits characteristics consistent with existing biomechanical understanding in terms of peak velocity, release angle range, the sequence of momentum transfer between segments, and local oscillation control. This suggests that the minimum-jerk constraint does not merely serve as a “mathematical smoothing” function, but likely captures, to some extent, the natural, economical, and repeatable control patterns inherent in dart throwing. It is worth noting that the “reasonableness” referred to here does not imply that the model has fully reconstructs the complete neuromuscular mechanisms of the movement, but rather that the generated results exhibit a high degree of consistency with the efficient organization of human movement at the kinematic level. Therefore, this study interprets the model as a biomechanically inspired reference generation tool rather than a complete reconstruction of the actual movement control mechanisms.

In terms of training feedback, the Z-score-based personalized recommendation generation method proposed in this study demonstrates the potential of moving from “motion recognition” to “training decision support.” While many existing motion analysis systems can perform motion classification, phase identification, or technical scoring, they ultimately still require coaches to reinterpret the results, making it difficult for these systems to be truly integrated into the training feedback loop. By quantitatively comparing deviations between the current movement and the distribution of high-quality movements specific to the individual athlete, this study ensures that “abnormality” is no longer simply interpreted as a deviation from the group mean, but rather defined as the degree of deviation from the athlete’s own stable performance range. This design has two significant implications: first, it enhances the individual specificity of diagnostic results and reduces the risk of misjudgment caused by group heterogeneity; on the other hand, it allows the generation of recommendations to correspond more directly to specific feature deviations, thereby enhancing the practicality of the feedback. The case analysis of S6 demonstrates that the system can identify issues such as insufficient trunk stability, abnormal elbow displacement, and imbalance in speed control, and generate training recommendations tailored to these deficiencies, preliminarily indicating that this method possesses practical application value.

\subsection*{Limitations of the Study}
However, the findings of this study should be interpreted within appropriate limitations; the conclusions do not imply that the system is currently capable of fully replacing coaching expertise or addressing the full complexity of darts training. First, from a data perspective, although this study constructed a relatively rare and specialized dataset of darts movements, the overall sample primarily consists of a limited number of participants, with an uneven distribution between professional and non-professional athletes. This suggests that the current results are better suited as a validation of methodological feasibility rather than as a basis for strong generalizations regarding broader populations, different training stages, or distinct technical styles. Particularly in personalized modeling scenarios, an insufficient sample size may cause the “optimal reference trajectory” to be influenced to some extent by short-term state fluctuations or occasional high-scoring samples. To further enhance the model’s robustness in the future, it will be necessary to incorporate larger-scale, cross-level, and cross-cycle longitudinal data.

Second, regarding the feature system, the 18 features extracted in this study are primarily based on observable external kinematic representations. While this approach ensures good interpretability and engineering feasibility, it also means that these features only partially capture the underlying mechanisms influencing dart-throwing performance. Performance in darts is affected not only by movement structure but also by the combined effects of various factors, including visual search strategies, attentional control, psychological pressure, equipment parameters, and even the on-site environment. In other words, this study currently constructs a “kinematics-centric” analytical framework rather than a comprehensive performance model. Therefore, interpreting the system’s output as a complete explanation of all factors affecting hitting results may overstate its diagnostic capabilities. A more cautious statement would be that the system can identify a subset of movement-level deviations significantly correlated with throwing outcomes and provide quantifiable evidence for training interventions, but it does not yet encompass all sources of information relevant to darts performance.

Furthermore, although this study attempts to move beyond static template matching, the proposed method still fundamentally relies on “historical high-performance samples” as a reference. This implies that its evaluation logic remains based on the premise that “past performance represents the current optimal direction.” While this premise is reasonable in most training contexts, it carries certain risks. For example, when an athlete is undergoing a technical transition, recovering from an injury, or actively experimenting with new movement strategies, historical high-scoring samples may no longer represent the future optimal solution. In such cases, the system may be more inclined to encourage a return to old patterns, lacking sensitivity to the potential advantages of new strategies. Therefore, from a critical perspective, although the method in this study offers greater individual adaptability than static group templates, it still falls under the category of “conservative optimization based on existing excellent performance” and has not yet truly achieved a forward-looking assessment of technical innovation pathways. In the future, integrating reinforcement learning, Bayesian updates, or performance tracking over longer periods may further enhance the system’s adaptability to the process of movement evolution.

Furthermore, the suggestion generation mechanism employed in this study is primarily based on rule-based feature deviation mapping and expert corpus organization. While this approach ensures the stability, interpretability, and professional consistency of the suggestions, it also limits expressive flexibility and reduces adaptability to complex scenarios. In actual training, an athlete’s technical issues are often not caused by a single abnormal feature but may result from the combined effects of multiple deviations. For example, an abnormal release angle may stem from insufficient wrist control or may be a chain reaction triggered by trunk sway, elbow path deviation, or an imbalance in speed and rhythm. Although the current system can identify multiple deviations, its recommendation generation still primarily relies on item-by-item prompts, and its ability to address the “causal hierarchy among deviations” and the “priority intervention sequence” remains limited. In other words, the system is currently better at answering “where the deviation lies,” but there is significant room for improvement in explaining “why the deviation occurs” and determining “which correction should be prioritized.” Addressing these challenges is essential for developing more advanced intelligent coaching systems in the future.

Despite these limitations, this study holds clear practical significance. First, at the training implementation level, the system proposed offers a relatively low-cost, markerless, and scalable quantitative assistance solution for darts, demonstrating strong practical feasibility. Compared to expensive laboratory-grade motion capture systems, the combination of Kinect and optical cameras more closely resembles everyday training scenarios, helping to bring biomechanical analysis tools from the laboratory to the front lines of training. Second, at the training philosophy level, this study emphasizes “within-individual comparison” rather than “replication of group standards.” This approach is methodologically significant for high-precision sports with strong individual variability, such as darts, and may also provide a reference for personalized analysis of other goal-oriented movements, including archery, basketball shooting, and batting. Finally, regarding human-machine collaboration, this study demonstrates a viable approach for integrating data-driven feedback with coaches’ experiential knowledge: the system does not replace the coach but provides more objective, granular, and verifiable diagnostic evidence, thereby enhancing the transparency and consistency of training decisions.

\subsection*{Future Work}
Future research can be advanced in three key directions. First, at the data level, the sample size should be expanded and a longitudinal tracking design introduced to validate the system’s robustness and generalization capabilities across different training stages, competitive levels, and movement styles. Second, at the perception level, multimodal signals such as eye tracking, electromyography, and heart rate variability should be integrated to reveal a more comprehensive coupling mechanism between “movement–perception–psychology” in dart-throwing performance. Third, at the decision-making level, develop feedback generation methods with stronger causal modeling capabilities and situational adaptability, enabling the system not only to identify movement deviations but also to make more precise inferences regarding the sources of these deviations, intervention priorities, and potential training benefits. Only by further deepening these aspects can a data-driven darts training assistance system evolve from a “usable motion analysis tool” into a “trusted intelligent training partner.”

Overall, the significance of this study should not be limited to the proposal of two specific algorithmic modules, but rather lies in establishing a new research and application paradigm for darts training: achieving personalized reference modeling and closed-loop feedback through data-driven methods while retaining biomechanical interpretability. This paradigm addresses the issues of high subjectivity and insufficient reproducibility in traditional empirical teaching, while avoiding the excessive disregard for individual differences inherent in static template matching. Although it remains in the stages of method validation and system refinement, this study provides insightful evidence and a pathway toward quantitative training support in high-precision, goal-oriented sports.

\section*{Conclusions}\label{sec5}
This study addresses three long-standing key issues in darts training—the scarcity of high-quality kinematic data, the reliance on static templates for movement evaluation, and the over-reliance on subjective experience for training feedback—by proposing and validating a data-driven training assistance system based on skeletal motion analysis and machine learning. Unlike traditional teaching methods that rely primarily on empirical observation, this study attempts to transform dart throwing—a goal-oriented movement characterized by short duration, high precision, and significant individual variation—into a quantifiable, modelable, and feedback-driven training decision-making problem, thereby providing methodological support for the scientific training of dart throwing. To achieve this goal, this paper first systematically identifies four core constraint dimensions of dart-throwing performance from a biomechanical perspective: the three-link coordination mechanism, release velocity, multi-joint angular configuration, and postural stability. Based on these, an analytical framework comprising 18 interpretable kinematic features was constructed. At the data level, the study integrated Kinect 2.0 depth sensors with optical cameras to collect 2,396 sets of throwing samples under markerless conditions that closely mimic real-world training environments, establishing a foundational dataset of dart-throwing motions encompassing both professional and non-professional athletes. At the model level, this paper further implemented two core functional modules: first, generating individualized optimal throwing reference trajectories based on the Minimum-jerk Principle and historical high-quality samples; second, quantifying deviations in single throwing motions and generating targeted training recommendations based on a Z-score mechanism and graded diagnostic logic. Experiments and case analyses demonstrate that the system can effectively identify key issues in movements and provide personalized feedback with a degree of biomechanical interpretation.

From a research perspective, the contribution of this paper lies not only in the development of a functional training assistance system, but also in the proposal of an approach to motion evaluation that differs from traditional static template matching. This paper emphasizes that high-level darts performance does not stem from the mechanical replication of a single “standard movement,” but rather manifests as a stable and effective coordination structure formed by the individual within the constraints of their own physical condition, movement habits, and task requirements. Based on this understanding, the reference trajectory generation method proposed in this paper no longer relies on a uniform group template as the sole criterion, but instead establishes an individualized reference frame based on the athlete’s own historical high-quality performance. This “within-individual reference” approach is theoretically more aligned with the functional equivalence and beneficial variability inherent in the learning of fine motor skills. It also shifts the focus of training feedback from “how much one deviates from the group standard” to “how much one deviates from one’s own optimal control range,” thereby enhancing the individual relevance of evaluation results and the practicality of training. At the same time, this paper demonstrates that biomechanical interpretation and data-driven modeling are not mutually exclusive but can form a complementary relationship within training support systems. Relying solely on traditional biomechanical research often fails to meet the demands for high-frequency, continuous, and individualized training feedback; conversely, relying solely on machine learning models without clear biomechanical semantic support tends to undermine the interpretability of system results and their acceptability to coaches. By designing features based on four core biomechanical dimensions and embedding them into trajectory fitting and diagnostic feedback processes, this study essentially explores a research pathway that “organizes problems through biomechanics and drives applications through data.” This approach also holds significant methodological transfer value for other high-precision, target-oriented sports beyond darts, such as archery, basketball shooting, and batting.

However, it should be cautiously noted that the conclusions of this paper primarily serve to validate the feasibility of the methodology rather than provide a definitive explanation of the principles underlying darts training. First, although the current data sample holds some exploratory value for darts research, its generalizability remains to be further tested in a larger and more diverse population due to limitations in the number of participants, the distribution of training levels, and the data collection settings. Second, the feature system constructed in this study primarily focuses on overt kinematic variables and has not yet incorporated factors equally important to dart performance, such as visual search, psychological pressure, equipment parameters, and environmental disturbances. Therefore, the system’s output should be understood as a quantitative representation of “movement-level deviations” rather than a complete explanation of all causes of hitting outcomes. Third, although the personalized reference model proposed in this study is more adaptable than static templates, it essentially still relies on historical high-quality motion samples. Therefore, its applicability for athletes undergoing technical transitions, motion reconstruction, or the development of innovative strategies requires further discussion.

Based on the above findings, future research could be deepened in three directions. First, expanding the sample size and incorporating a longitudinal tracking design to enhance the model’s ability to adapt to training evolution, individual differences, and groups at varying competitive levels. Second, integrating multimodal data—such as eye-tracking, electromyography, and heart rate variability—to construct a more comprehensive “motion-perception-psychology” coupling analysis framework, thereby improving the system’s depth of interpretation regarding throwing performance. Third, developing feedback mechanisms with greater causal inference capabilities and situational adaptability, enabling the system not only to identify movement deviations but also to further determine the source of deviations, intervention priorities, and assess potential training benefits, thereby driving the evolution of assisted training systems from mere “motion analysis tools” to “intelligent decision support systems.”

Overall, this paper provides a data-driven training support solution for darts that balances biomechanical interpretability, the preservation of individual differences, and engineering feasibility. This study not only establishes a closed-loop system at the technical level—from motion capture to personalized feedback—but also advances a conceptual shift in darts training from an experience-driven approach toward quantitative, dynamic, and individualized analysis. Although the system still requires continuous refinement through richer data, additional modality signals, and more complex training scenarios, the findings of this study indicate that data-driven methods hold significant promise for providing a more scientific, transparent, and scalable approach to supporting training in high-precision, goal-oriented sports.

\section*{Acknowledgements}
The authors would like to thank all the members of the research team for their valuable discussions and support throughout this study. The authors also gratefully acknowledge the support provided by Qinghai University, Chongqing University of Technology, the University of Macau, The Hong Kong Polytechnic University, and Xiaoping Technology Innovation Lab, Zhongshan Xiaolan Senior High School.

\section*{Author contributions statement}
Z.C. and D.H. contributed equally to this work. 
\textbf{Z.C.:} Conceptualization, Methodology, Investigation, Data curation, Formal analysis, Writing--original draft. 
\textbf{D.H.:} Conceptualization, Methodology, Investigation, Formal analysis, Writing--review and editing. 
\textbf{J.F.:} Conceptualization, Supervision, Project administration, Resources. 
\textbf{X.C.:} Data curation. 
\textbf{Y.L.:} Data curation. 
\textbf{X.Z.:} Data curation. 
\textbf{X.H.:} Conceptualization, Supervision, Resources. 
\textbf{All authors:} Read and approved the final manuscript.

\section*{Competing interests}
The authors declare no competing interests.

\section*{Data Availability}
The data that support the findings of this study are available from the corresponding author upon reasonable request.

\section*{Funding}
No Funding.

\bibliography{sample}

\end{document}